\documentclass{article} 
\usepackage{iclr2027_conference,times}

\usepackage{amsmath,amsfonts,bm}

\def\eqref#1{equation~\ref{#1}}

\def\1{\bm{1}}

\def\vx{{\bm{x}}}
\def\vy{{\bm{y}}}

\DeclareMathAlphabet{\mathsfit}{\encodingdefault}{\sfdefault}{m}{sl}
\SetMathAlphabet{\mathsfit}{bold}{\encodingdefault}{\sfdefault}{bx}{n}

\usepackage[T1]{fontenc}
\usepackage{hyperref}
\usepackage{url}
\usepackage{graphicx}
\usepackage{booktabs}
\usepackage{multirow}
\usepackage{amsmath}
\usepackage{bm}
\usepackage{tikz}
\usetikzlibrary{arrows.meta,backgrounds,calc,fit,positioning}

\title{CryoTracer: A Synthetic-to-Real System for\\
Filament Instance Tracing in Cryo-EM}

\author{%
  Radosław Kuczbański$^{1*}$,
  Wojciech Kozłowski$^{1}$,
  Mateusz Grzesiuk$^{1}$, \\
  \textbf{Konrad Karanowski}$^{1}$,
  \textbf{Lukas Frey}$^{2}$,
  \textbf{Andrzej Rzepiela}$^{2}$,
  \textbf{Maciej Zięba}$^{1,3}$ \\
  $^{1}$WUST, Poland $\quad$ $^{2}$ETH Zurich, Switzerland $\quad$ $^{3}$Tooploox, Poland \\
  *\texttt{radoslaw.kuczbanski@pwr.edu.pl} \\
}

\iclrfinalcopy 
\begin{document}

\maketitle
\lhead{Preprint}

\begin{abstract}

Particle picking in cryo-electron microscopy is commonly formulated as an object-detection problem in which individual particles are represented by bounding boxes, a natural representation for globular particles but a poor match for filaments that can extend across an entire micrograph. Existing methods reconstruct filament trajectories by linking local predictions in a separate post-processing stage, decoupling detection from instance formation and creating potential ambiguities at gaps and crossings. Directly preserving filament identity is particularly useful for polymorphic specimens, where segments originating from the same filament should remain associated. However, learning such instance-level representations is challenging because expert centerline annotations are scarce and costly to obtain. We introduce CryoTracer, a synthetic-to-real framework that addresses this limitation through procedural micrograph generation, transformer-based polyline prediction, and specimen-specific fine-tuning. Procedural pretraining exposes the model to diverse filament geometries, intersections, and imaging conditions before adaptation using a limited number of fully annotated experimental micrographs. CryoTracer represents filaments as ordered centerline polylines and optimizes their geometry using an orientation-invariant discrete Fréchet objective, enabling consistent learning regardless of tracing direction. Evaluation on established and newly expert-annotated datasets shows that CryoTracer improves instance-level filament recovery over existing particle-detection approaches. These results establish end-to-end polyline prediction as a practical alternative to detection-and-linking pipelines for recovering complete filament instances with limited micrograph annotation budgets. Code and data are available at \url{https://github.com/cryotracer/cryotracer}.

\end{abstract}

\section{Introduction}

Cryo-electron microscopy (cryo-EM) uses two-dimensional images of vitrified specimens to determine three-dimensional macromolecular structures. Software packages such as RELION~\cite{Scheres2012} and cryoSPARC~\cite{punjani2017cryosparc} require an initial particle-picking step to identify particles in noisy micrographs. For filamentous specimens, this step must recover centerlines and distinguish individual filaments. Helical reconstruction workflows extract overlapping segments along these traces. Grouping segments by their parent filament allows an entire filament to be inspected or excluded and classification evidence to be combined across its segments. This grouping is particularly useful for specimens containing multiple structural polymorphs that must be separated before reconstruction~\cite{frey2024polymorphism}. For example, $\alpha$-synuclein is a major component of the protein aggregates associated with Parkinson's disease and dementia with Lewy bodies~\cite{Spillantini1997,Schweighauser2020,Yang2022,Enomoto2025}. Its fibrils can occur in distinct structural polymorphs~\cite{li2018polymorphs}, and filament-level subset selection has been used when processing mixed $\alpha$-synuclein polymorph datasets~\cite{Lvestam2023}. Preserving filament identity during picking requires following each filament through curves, crossings, and low-contrast regions. A fading signal can resemble an endpoint, while a crossing can leave several possible continuations. A tracer must consider the surrounding filament geometry to avoid splitting one filament into fragments or joining different filaments. Preparing real training data requires annotators to mark each filament's endpoints and place points along its centerline. At crossings or gaps in the visible signal, they must decide which segments belong to the same filament.

Established particle pickers such as crYOLO and Topaz assemble filament traces from local predictions. crYOLO's filament mode requires training on manually annotated filaments~\cite{wagner2020two}. It connects local bounding boxes into traces without directly predicting a complete ordered centerline. Topaz's filament extension~\cite{lovestam2022highthroughput} selects regions above a threshold in a heatmap of likely particle locations and reduces them to thin lines. It then detects straight segments and merges nearby segments with similar orientations into longer traces. In both workflows, filament continuity is reconstructed through geometric rules with user-selected parameters, which required expert tuning in our comparisons. On $\alpha$-synuclein, both methods produced more numerous, shorter traces than the annotated filaments, a pattern consistent with fragmentation despite substantial spatial coverage.

To address these limitations, we introduce CryoTracer, a synthetic-to-real system that predicts each filament as an ordered five-point centerline from a full micrograph. Direct centerline prediction makes filament continuity part of the learning task, removing the need to assemble traces from local detections using geometric linking rules. The model adapts RF-DETR~\cite{robinson2026rfdetrneuralarchitecturesearch} with curve proposals refined through deformable attention and successive decoder stages. An orientation-invariant discrete Fr\'echet objective supervises curve matching and regression. To reduce reliance on expert annotations, procedural micrograph generation provides centerlines, endpoints, and filament identities by construction, while varying curvature, density, crossings, width, and background appearance. This enables pretraining on diverse examples without tracing each micrograph manually. Fine-tuning on real annotations then adapts the model to the specimen and acquisition conditions.

Our contributions are as follows.
\begin{itemize}
    \item We introduce a synthetic-to-real filament-tracing system that learns filament continuity through direct centerline prediction and reduces reliance on expert annotations through procedural pretraining followed by specimen-specific fine-tuning.
    \item We release filament annotations for EMPIAR-10022 and evaluate on an in-house $\alpha$-synuclein dataset comprising 325 expert-annotated micrographs from 11 reconstruction sessions.
    \item We compare CryoTracer with established cryo-EM methods under standard and limited-annotation regimes, using instance-level and spatial-overlap metrics to distinguish complete filament recovery from local coverage.
\end{itemize}

\section{Related work}

\paragraph{Set detection and centerline representations.}
DETR~\cite{carion2020end} formulates object detection as prediction of a set
of instances with bipartite matching. RF-DETR builds on transformer detection
with an architecture designed for efficient inference
\cite{robinson2026rfdetrneuralarchitecturesearch}. CryoTracer adapts this
detector to ordered curves. Centerline representations themselves are not
unique to this work: Alonso and Kirkegaard~\cite{alonso2023fast} predict
centerlines of slender bodies in microscopy videos using a convolutional
network trained on synthetic data. Their method uses temporal context and
targets detection and tracking of moving organisms. CryoTracer instead
conditions a transformer decoder on polyline references in individual
cryo-EM micrographs and fine-tunes on specimen-specific annotations.

\paragraph{Particle picking in cryo-EM.}
Topaz~\cite{Bepler2019} uses positive-unlabeled learning to train particle
pickers with sparse annotations. crYOLO's filament mode~\cite{wagner2020two}
combines learned box detections with tracing post-processing. Other learned
particle pickers include EPicker~\cite{zhang2022epicker}, which uses
CenterNet and continual learning, and CryoTransformer~\cite{Dhakal2024},
which adapts transformer detection to particle picking. These methods
illustrate the range of learned particle-localization approaches. Their
particle-level outputs do not directly provide the ordered sequences of points
along individual filaments considered here.

\paragraph{Filament tracing.}
FibrilFinder~\cite{thurber2021automated} uses classical image processing to
identify amyloid fibrils and prepare coordinates for helical reconstruction.
RELION 5.1's AmyTracer~\cite{lovestam2026relion} combines the characteristic
$4.75$\,\AA{} repeat signal of amyloid cross-$\beta$ sheets with U-Net
segmentation. This repeat reflects the spacing of $\beta$-strands along the
filament axis and produces a directional Fourier signal. Filament centerlines
are then extracted by thresholding and skeletonizing the predicted probability
maps.

\section{Method}
\label{sec:method}

\begin{figure}[t]
\centering
\begin{tikzpicture}[
    x=1mm,y=1mm,
    font=\sffamily\fontsize{8}{9.5}\selectfont,
    heading/.style={font=\sffamily\bfseries\fontsize{8.5}{10}\selectfont},
    note/.style={font=\sffamily\fontsize{7}{8.5}\selectfont,align=center},
    flow/.style={-{Latex[length=1.5mm,width=1.1mm]},draw=black!65,
        line width=0.7pt,line cap=round,line join=round},
    frame/.style={draw=black!25,fill=white,line width=0.45pt},
    reference/.style={draw=ctReference,line width=1.05pt,line join=round},
    refined/.style={draw=ctRefined,line width=1.05pt,line join=round},
    sample/.style={draw=ctSample,fill=ctSample!75,line width=0.45pt}
]
\definecolor{ctReference}{HTML}{C46B08}
\definecolor{ctSample}{HTML}{8048B4}
\definecolor{ctRefined}{HTML}{008CA8}
\def\ctReferencePath{(5,3)--(9,8)--(10,18)--(19,16)--(24,23)}
\def\ctFeaturePath{(4,7)--(7,16)--(16,17)--(24,21)--(27,28)}
\path[use as bounding box] (0,-2) rectangle (139.7,73);

\node[heading] at (16,69) {(a) Proposals};
\node[heading] at (67,69) {(b) Attention};
\node[heading] at (121,69) {(c) Refinement};
\node[note] at (16,63) {Top-$Q$ encoder curves};
\node[note] at (67,63) {Four feature levels};
\node[note] at (121,63) {Update all $K$ points};

\draw[frame] (1,26) rectangle (31,56);
\begin{scope}[shift={(1,26)}]
    \foreach \curve in {
        {(2,19)--(4,22)--(4,25)--(7,27)--(10,28)},
        {(10,25)--(14,27)--(18,28)--(22,27)--(26,29)},
        {(20,3)--(23,5)--(25,8)--(26,11)--(28,14)},
        {(2,2)--(7,3)--(11,2)--(15,4)--(18,4)},
        {(2,9)--(3,12)--(2,15)--(4,17)--(6,18)},
        {(15,7)--(18,9)--(21,9)--(23,12)--(27,13)},
        {(16,22)--(19,23)--(22,22)--(26,23)--(28,25)}
    } \draw[ctReference!40,line width=0.7pt,line join=round] \curve;
    \draw[reference] \ctReferencePath;
    \foreach \x/\y in {5/3,9/8,10/18,19/16,24/23}
        \fill[ctReference] (\x,\y) circle[radius=0.65mm];
\end{scope}
\node[note] at (16,20) {$K=5$ points per curve\\$Q=50$ selected curves};

\foreach \d in {3,2,1,0}
    \draw[frame] (52+\d,26+\d) rectangle (82+\d,56+\d);
\begin{scope}[shift={(52,26)}]
    \clip (0,0) rectangle (30,30);
    \draw[step=5,black!10,line width=0.25pt] (0,0) grid (30,30);
    \draw[ctRefined!9,line width=2.6mm,line cap=round,line join=round]
        \ctFeaturePath;
    \draw[ctRefined!16,line width=1.2mm,line cap=round,line join=round]
        \ctFeaturePath;
    \foreach \vx/\vy/\sx/\sy/\radius in {
        5/3/4.5/6.5/0.85,5/3/7/2/0.35,
        9/8/7/14.5/0.95,9/8/12/9/0.4,
        10/18/15.5/17/1.05,10/18/8/20.5/0.35,
        19/16/22/20/0.9,19/16/22/15/0.4,
        24/23/26.5/27/0.85,24/23/27/21/0.35} {
        \draw[ctSample!75,line width=0.5pt] (\vx,\vy) -- (\sx,\sy);
        \draw[sample] (\sx,\sy) circle[radius=\radius mm];
    }
    \draw[reference] \ctReferencePath;
    \foreach \x/\y in {5/3,9/8,10/18,19/16,24/23}
        \fill[ctReference] (\x,\y) circle[radius=0.65mm];
\end{scope}
\node[note] at (67,20) {Sample around each point\\Aggregate weighted features};

\draw[frame] (106,26) rectangle (136,56);
\begin{scope}[shift={(106,26)}]
    \draw[ctRefined!9,line width=2.6mm,line cap=round,line join=round]
        \ctFeaturePath;
    \draw[ctRefined!16,line width=1.2mm,line cap=round,line join=round]
        \ctFeaturePath;
    \draw[reference,dashed,opacity=0.65] \ctReferencePath;
    \foreach \x/\y in {5/3,9/8,10/18,19/16,24/23}
        \draw[ctReference,fill=white,line width=0.6pt]
            (\x,\y) circle[radius=0.55mm];
    \draw[refined] (4.5,5.5)--(7.5,13)--(14.5,17)--(22,19)--(26,26);
    \foreach \x/\y in {4.5/5.5,7.5/13,14.5/17,22/19,26/26}
        \fill[ctRefined] (\x,\y) circle[radius=0.65mm];
\end{scope}
\node[note] at (121,20) {Predict coordinate residuals\\Refine the reference curve};

\draw[flow] (33,41) -- (50,41);
\draw[flow] (87,41) -- (104,41);
\draw[flow,draw=ctRefined,rounded corners=1.5mm]
    (104,32) -- (97,32) -- (97,9) -- (44,9) -- (44,32) -- (50,32);
\node[note,fill=white,inner sep=1mm,text=ctRefined] at (70.5,9)
    {Updated reference\\Repeat for six decoder layers};

\draw[reference] (1,0) -- (6,0);
\fill[ctReference] (3.5,0) circle[radius=0.55mm];
\node[note,anchor=west] at (7,0) {Reference curve};
\draw[sample] (49,0) circle[radius=0.4mm];
\draw[sample] (52,0) circle[radius=0.8mm];
\node[note,anchor=west] at (54,0) {Samples: area $\propto$ weight};
\draw[refined] (106,0) -- (111,0);
\fill[ctRefined] (108.5,0) circle[radius=0.55mm];
\node[note,anchor=west] at (112,0) {Refined curve};
\end{tikzpicture}
\caption{\textbf{Polyline prediction and iterative refinement in CryoTracer.}
(a) The $Q=50$ highest-scoring encoder proposals provide initial reference polylines with $K=5$ ordered points.
(b) Cross-attention updates each query using encoder features around its
current polyline (orange). Purple markers indicate sampling locations,
with marker area proportional to attention weight.
(c) The updated query refines the polyline, and the resulting curve (teal)
guides attention in the next decoder layer. This process progressively
refines the predictions.}
\label{fig:polyline-decoding}
\end{figure}
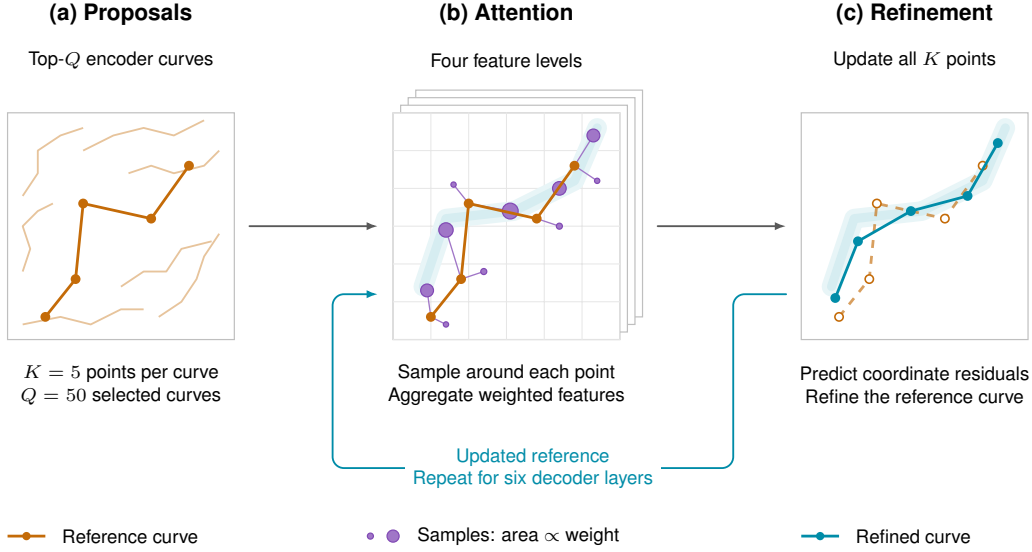

Below, we present \emph{CryoTracer}, a method for detecting filament instances in high-resolution, noisy micrographs. First, we describe our proposed architecture, which modifies RF-DETR to enable a polyline-based representation of detected structures. Next, we explain a two-stage training paradigm that addresses the problem of limited annotated real data. The key components of our method are illustrated in Figure~\ref{fig:polyline-decoding}.

\paragraph{Architecture overview.}
Our detection model consists of an encoder $\mathcal{E}_\theta$ and a decoder $\mathcal{D}_\phi$, parameterized by $\theta$ and $\phi$, respectively. The decoder contains several learnable projection modules $(\psi, \Phi, \Gamma)$. Although these modules have separate parameter subsets, we denote all of them by $\phi$ for simplicity, emphasizing that they jointly constitute the decoder $\mathcal{D}_\phi$.

\paragraph{Feature extraction.}
Given an input micrograph $\mathbf{X} \in \mathbb{R}^{1 \times W \times H}$, the encoder $\mathcal{E}_\theta$ extracts a set of feature maps $\{\mathbf{F}_l\}_{l=1}^L$ from $L$ selected intermediate layers. Each spatial location of each feature map $\mathbf{F}_l$ defines one candidate polyline proposal.

\paragraph{Initial proposal construction.}
For a spatial location $(x,y)$ in feature map $\mathbf{F}_l$, we obtain a latent proposal representation
\begin{equation}
\mathbf{z}_i = \psi_\phi(\mathbf{F}_{l,x,y}),
\end{equation}
where, for simplicity, we use a single index $i$ to enumerate proposals across all feature maps. Each proposal is additionally associated with a center $\mathbf{c}_i$, corresponding to the normalized spatial position $(x,y)$, and a scale $\mathbf{r}_i$ determined by the feature-map level $l$. Coarser feature maps correspond to larger spatial scales.

From the latent representation $\mathbf{z}_i$, we predict a confidence score $\gamma_i = \Phi_\phi(\mathbf{z}_i)$.
We retain the $Q=50$ proposals with the highest confidence scores. This exceeds the observed filament counts in both datasets (means: 20.3 and 5.9), while limiting decoder cost. Given the image sizes and filament widths, substantially denser scenes would also increase overlap and tracing ambiguity. For each retained proposal, an MLP $\Gamma_\phi$ predicts offsets for $K$ ordered polyline points:
\begin{equation}
\mathbf{D}_i = \Gamma_\phi(\mathbf{z}_i),
\qquad
\mathbf{D}_i \in \mathbb{R}^{K \times 2}.
\end{equation}
The initial polyline is then constructed as
\begin{equation}
\mathbf{P}_i =
\left(
\mathbf{c}_i
+ \tanh(\mathbf{D}_{i,k}) \odot \mathbf{r}_i
\right)_{k=1}^{K}.
\end{equation}
Each row $\mathbf{D}_{i,k}$ specifies the offset of the $k$-th polyline point relative to the proposal center. The $\tanh$ function bounds the predicted offsets, while $\mathbf{r}_i$ scales them according to the spatial scale associated with the corresponding feature-map level. The resulting $K$ ordered points form the initial polyline prediction $\mathbf{P}_i$.

\paragraph{Iterative polyline refinement.}
Each initial polyline is predicted independently from a single feature-map location $\mathbf{F}_{l,x,y}$. Consequently, the initial predictions have two limitations. First, multiple proposals may correspond to the same filament, while other filaments may receive no proposal. Second, predicting the entire polyline from a single feature representation may result in inaccurate localization or shape estimates. Following RF-DETR, we address both limitations through an iterative refinement procedure depicted in Figure~\ref{fig:polyline-decoding} b) and c).

For each decoder query slot $i$, we initialize the query content $\mathbf{q}_i^{(0)}$ from a learned query embedding shared across images. The selected encoder proposal $\mathbf{P}_i$ supplies the initial reference geometry, $\mathbf{P}_i^{(0)}=\mathbf{P}_i$.

At decoder layer $t$, the queries first interact through self-attention,
\begin{equation}
\tilde{\mathbf{q}}_i^{(t)}
=
\operatorname{SelfAttn}_\phi\left(
\mathbf{q}_i^{(t-1)},
\{\mathbf{q}_j^{(t-1)}\}_{j=1}^{Q}
\right),
\end{equation}
allowing information to be exchanged between proposals. This interaction helps the decoder reason jointly about the $Q$ candidate objects rather than refining each proposal independently.

Next, cross-attention aggregates information from all encoder feature levels $\{\mathbf{F}_l\}_{l=1}^L$ into query $\tilde{\mathbf{q}}_i^{(t)}$. The current polyline $\mathbf{P}_i^{(t-1)}$ provides the reference locations for sampling features around each of its points:
\begin{equation}
\mathbf{q}_i^{(t)}
=
\operatorname{CrossAttn}_\phi\left(
\tilde{\mathbf{q}}_i^{(t)},
\{\mathbf{F}_l\}_{l=1}^L;
\mathbf{P}_i^{(t-1)}
\right).
\end{equation}
Consequently, the query is updated using features sampled around its current polyline across all feature levels.

Finally, the updated query predicts a correction to the current polyline,
\begin{equation}
\mathbf{P}_i^{(t)}
=
\operatorname{Refine}_\phi
\left(
\mathbf{P}_i^{(t-1)},\mathbf{q}_i^{(t)}
\right).
\end{equation}
The refined polyline $\mathbf{P}_i^{(t)}$ and query $\mathbf{q}_i^{(t)}$ are passed to the next decoder layer, progressively improving the predictions. After the final layer, the resulting polylines $\{\hat{\mathbf{P}}_i\}_{i=1}^Q$ form the decoder outputs.

\paragraph{Training objective.}
Since we are representing objects with polylines, a simple regression from off-the-shelf DETR methods is not applicable. Instead, to compare the predicted polyline $\hat{\mathbf{P}}_i$ with a ground truth polyline $\mathbf{Y}_j$, we use the Fréchet Distance $d_F(\hat{\mathbf{P}}_i, \mathbf{Y}_j)$, which measures the spatial distance between two polylines rather than exact point-to-point alignment. To make the distance invariant to point direction (forward vs backward), we define it as 
\begin{equation}
    \bar{d}_F(\hat{\mathbf{P}}_i, \mathbf{Y}_j) = \min \{ d_F(\hat{\mathbf{P}}_i, \mathbf{Y}_j), d_F(\text{rev}(\hat{\mathbf{P}}_i), \mathbf{Y}_j)\},
    \label{eq:oriented-frechet}
\end{equation}
where $\text{rev}$ reverses the order of points. To find a proper match between the sets of predictions $\{\hat{\mathbf{P}}_i\}_{i=1}^Q$ and ground truths $\{\mathbf{Y}_j\}_{j=1}^N$, we use Hungarian matching to find a minimum-cost one-to-one assignment that balances geometric distance $\bar{d}_F$ and prediction confidence. Predictions not selected by the assignment are optimized as no-object (polyline confidence $\gamma$ is minimized for them via the cross entropy loss term $\mathcal{L}_{CE}$). For matched polylines, we optimize the following criterion
\begin{equation}
    \mathcal L_{F}
    =
    \frac{1}{\max(N,1)}
    \sum_{j=1}^{N}
        \bar{d}_{F}
        \left(
            \hat{\mathbf P}_{\pi(j)},
            \mathbf Y_j
        \right),
\end{equation}
where $\pi(j)$ is the index of the prediction matched to the $j$-th ground truth polyline. The cross-entropy loss $\mathcal{L}_{CE}$ also encourages high confidence for matched predictions. Combining classification and geometry gives
\begin{equation}
    \mathcal{L} = \mathcal{L}_{CE} + \lambda_F \mathcal{L}_{F},
    \label{eq:prediction-loss}
\end{equation}
with geometric loss weight $\lambda_F=10$.

We apply Equation~\ref{eq:prediction-loss} independently to the final decoder predictions and the $Q$ selected initial encoder proposals, obtaining $\mathcal{L}_{\mathrm{dec}}$ and $\mathcal{L}_{\mathrm{enc}}$, respectively. Each prediction set has its own Hungarian assignment. Encoder supervision directly trains the proposal heads because the initial coordinates are detached before decoder refinement. We therefore compute $\mathcal{L}_{\mathrm{enc}}$ from an undetached copy of the selected proposals.

The decoder and encoder losses are combined as
\begin{equation}
    \mathcal{L}_{\mathrm{total}}
    = \lambda_{\mathrm{dec}}\mathcal{L}_{\mathrm{dec}}
    + \lambda_{\mathrm{enc}}\mathcal{L}_{\mathrm{enc}},
\end{equation}
where $\lambda_{\mathrm{dec}}=\lambda_{\mathrm{enc}}=1$.

\paragraph{Synthetic pretraining.}
Starting from RF-DETR weights, we pretrain on procedural micrographs with
exact centerline labels. The generator varies filament geometry, crossings,
density, and image appearance, and includes empty fields. Procedural
generation supplies diverse examples without additional manual tracing,
providing supervision for learning filament shape and instance separation
before adaptation to real data.

\paragraph{Real data finetuning.}
The synthetic checkpoint initializes finetuning on annotated real
micrographs. This stage uses the available centerline annotations to adapt
the pretrained representation to specimen-specific filament morphology and
experimental image appearance. Both stages use the same architecture and
objective, with all parameters trainable. Data generation and preprocessing
are specified in Appendix~\ref{sec:appendix-data}; optimization and checkpoint
selection are specified in Appendix~\ref{sec:implementation-details}.

\paragraph{Inference and non-maximum suppression.}
At inference, we retain predictions with confidence at least $0.5$ and apply
polyline non-maximum suppression (NMS) to remove duplicate filament detections.
Duplicates are identified by similar unoriented directions, substantial
projected overlap, and small perpendicular separation, preserving the
geometry of the higher-confidence curve. We use the same thresholds across
both datasets and evaluation regimes (Appendix~\ref{sec:inference-export}).

\section{Experiments}
\label{sec-experiments}

We evaluate how accurately CryoTracer and established cryo-EM methods recover
complete filaments on EMPIAR-10022 and $\alpha$-synuclein. We use a
70/15/15\% train/validation/test split in the standard regime.

A typical reconstruction session contains about 2,000 micrographs, but
experts may annotate only 30 to 50 to adapt a tracing model to a new specimen.
We study this limited-annotation setting with a practical regime, using a
15/15/70\% train/validation/test split.

We report precision, recall, and F1 for filament detection, with Box IoU and
coverage as complementary overlap metrics.
We detail the splits, baseline configurations, and prediction conversions
in Appendix~\ref{sec:experimental-protocol}.

\subsection{Evaluation metrics}
\label{sec:evaluation-metrics}

We measure complete-instance recovery using orientation-invariant discrete
Fr\'echet matching between predicted and annotated centerlines. Precision,
recall, and F1 are computed per micrograph and averaged across the evaluation
set. Matching is sensitive to fragmentation and endpoint errors. Our primary
metric, mF1, averages these
macro F1 scores equally over distance tolerances of
$25$, $50$, $100$, $150$, and $200$\,\AA{}. We compute mRecall and mPrecision
by the same averaging procedure. Main results summarize five training seeds
for confirmation, while ablations use only one seed. Appendix tables give
per-tolerance scores and sample standard deviations.

Box IoU and coverage provide complementary measures of spatial agreement
between predicted and annotated filament regions. Pick counts and mean pick
lengths in nanometres describe predictions and ground truth on the same
evaluation images. Appendix~\ref{sec:evaluation-metric-details} specifies
matching, resampling, and overlap definitions.

\subsection{Main results}
\label{sec:main-results}

CryoTracer achieves the highest mF1 on both datasets and in both evaluation regimes
(Tables~\ref{tab:empiar10022-small} and~\ref{tab:asyn-small}).

\paragraph{EMPIAR-10022.}
CryoTracer achieves mF1 of $0.63$ and $0.54$ in the standard and practical
regimes, compared with $0.08$ and $0.09$ for Topaz, the strongest baseline
by mF1. CryoTracer leads both overlap metrics in the standard regime,
whereas Topaz leads them in the practical regime.

\begin{table}[!htbp]
\centering
\caption{Filament detection on EMPIAR-10022 in the standard and practical regimes. mF1, mRecall, and mPrecision average per-micrograph scores over five distance tolerances (Section~\ref{sec:evaluation-metrics}). Box IoU measures spatial overlap; coverage is the fraction of annotated area recovered. Pick count and mean polyline length (nm) summarize filament traces, with ground truth as a reference. CryoTracer values are means over five training seeds. Bold marks the highest unrounded score per metric and regime; pick statistics are descriptive.}
\label{tab:empiar10022-small}
\resizebox{\linewidth}{!}{%
\begin{tabular}{lrrrrrrr}
\toprule
Method & mF1 & mRecall & mPrecision & Box IoU & Coverage & Pick count & Mean pick length (nm) \\
\midrule
\multicolumn{6}{l}{\textit{Standard regime: ground truth}} & 347 & 181 \\
\cmidrule(lr){1-8}
CryoTracer (ours) & \textbf{0.63} & \textbf{0.63} & \textbf{0.63} & \textbf{0.84} & \textbf{0.90} & 337 & 176 \\
Topaz & 0.08 & 0.13 & 0.06 & 0.82 & 0.88 & 947 & 56 \\
crYOLO & 0.03 & 0.03 & 0.03 & 0.66 & 0.69 & 329 & 122 \\
\midrule
\multicolumn{6}{l}{\textit{Practical regime: ground truth}} & 1,578 & 190 \\
\cmidrule(lr){1-8}
CryoTracer (ours) & \textbf{0.54} & \textbf{0.53} & \textbf{0.56} & 0.81 & 0.86 & 1,474 & 184 \\
Topaz & 0.09 & 0.16 & 0.06 & \textbf{0.86} & \textbf{0.92} & 4,607 & 61 \\
crYOLO & 0.02 & 0.02 & 0.02 & 0.65 & 0.68 & 1,559 & 118 \\
\bottomrule
\end{tabular}%
}
\end{table}

\paragraph{$\alpha$-synuclein.}
CryoTracer achieves mF1 of $0.40$ and $0.34$ in the standard and practical
regimes, compared with $0.11$ and $0.09$ for the RELION
Amyloid model, the strongest baseline by mF1. CryoTracer leads Box IoU in
both regimes; Topaz has higher practical-regime coverage.

\begin{table}[!htbp]
\centering
\caption{Filament detection on $\alpha$-synuclein in the standard and practical regimes. mF1, mRecall, and mPrecision average per-micrograph scores over five distance tolerances (Section~\ref{sec:evaluation-metrics}). Box IoU measures spatial overlap; coverage is the fraction of annotated area recovered. Pick count and mean polyline length (nm) summarize filament traces, with ground truth as a reference. CryoTracer values are means over five training seeds. Bold marks the highest unrounded score per metric and regime; pick statistics are descriptive.}
\label{tab:asyn-small}
\resizebox{\linewidth}{!}{%
\begin{tabular}{lrrrrrrr}
\toprule
Method & mF1 & mRecall & mPrecision & Box IoU & Coverage & Pick count & Mean pick length (nm) \\
\midrule
\multicolumn{6}{l}{\textit{Standard regime: ground truth}} & 272 & 203 \\
\cmidrule(lr){1-8}
CryoTracer (ours) & \textbf{0.40} & \textbf{0.43} & \textbf{0.39} & \textbf{0.75} & \textbf{0.84} & 278 & 176 \\
Topaz & 0.01 & 0.04 & 0.01 & 0.59 & 0.80 & 1,659 & 33 \\
crYOLO & 0.01 & 0.01 & 0.01 & 0.60 & 0.66 & 540 & 60 \\
RELION Amyloid & 0.11 & 0.11 & 0.11 & 0.59 & 0.63 & 342 & 94 \\
\midrule
\multicolumn{6}{l}{\textit{Practical regime: ground truth}} & 1,364 & 203 \\
\cmidrule(lr){1-8}
CryoTracer (ours) & \textbf{0.34} & \textbf{0.36} & \textbf{0.33} & \textbf{0.69} & 0.80 & 1,374 & 181 \\
Topaz & 0.02 & 0.05 & 0.01 & 0.63 & \textbf{0.83} & 7,891 & 34 \\
crYOLO & 0.00 & 0.00 & 0.00 & 0.58 & 0.62 & 2,498 & 61 \\
RELION Amyloid & 0.09 & 0.09 & 0.09 & 0.61 & 0.66 & 1,914 & 88 \\
\bottomrule
\end{tabular}%
}
\end{table}

\paragraph{Implications for filament processing.}
On standard $\alpha$-synuclein, Topaz produces 1,659 picks with
mean pick length $33$\,nm, compared with 272 annotated filaments averaging
$203$\,nm. CryoTracer produces 278 picks averaging $176$\,nm. The shorter,
more numerous Topaz picks are consistent with fragmentation, although false
positives and endpoint errors can also contribute.
Thus, substantial spatial overlap can coexist with poor recovery of
individual filaments.
Figure~\ref{fig:pick-lengths} illustrates these
differences in the picked-filament length distributions.

\begin{figure*}[t]
    \centering
    \includegraphics[width=0.8\textwidth,height=0.82\textheight,keepaspectratio]{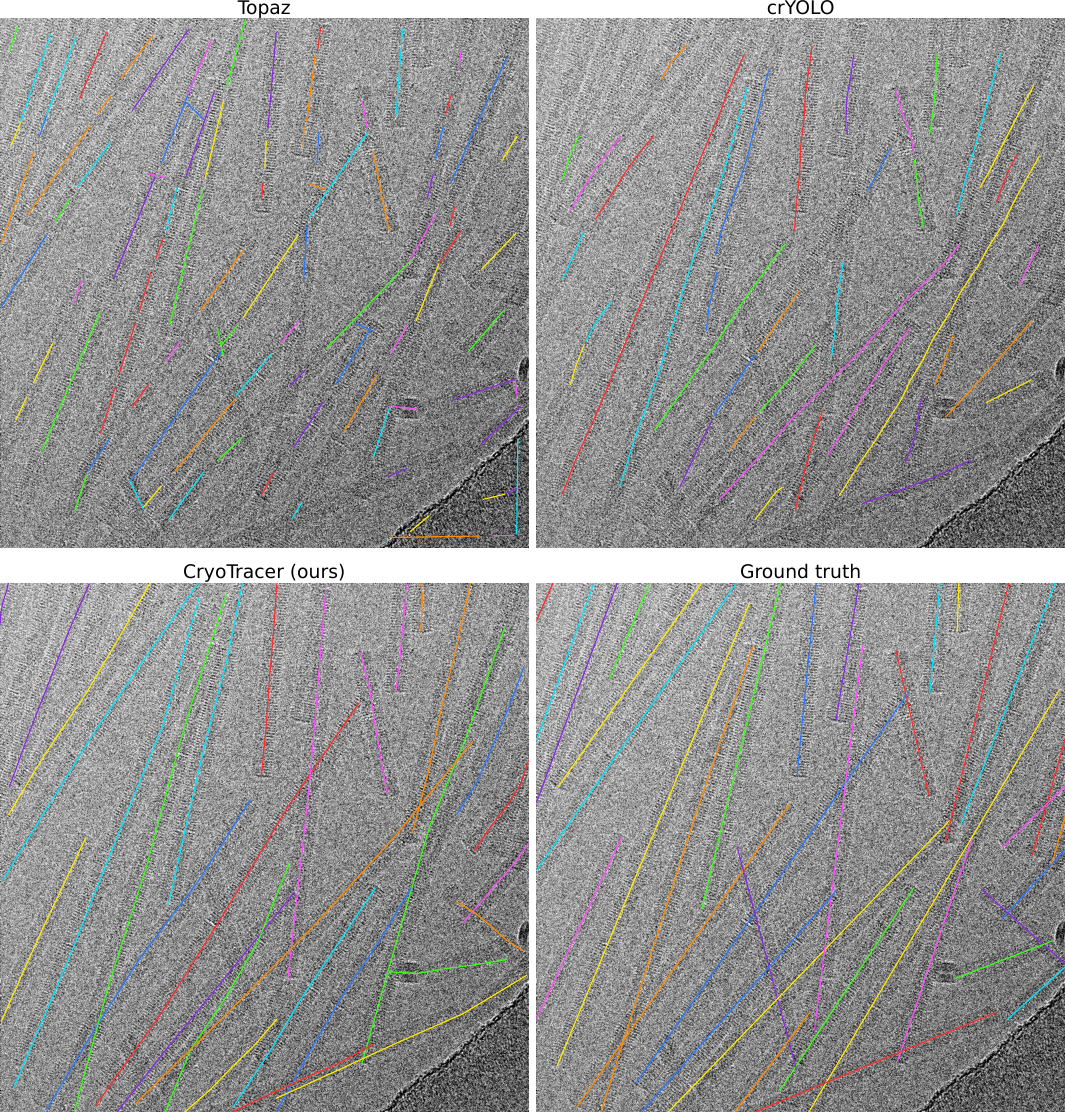}
    \caption{EMPIAR-10022, standard regime. One selected micrograph from the corresponding evaluation set, with method predictions and ground truth labeled in the panels. Images are low-pass filtered and contrast adjusted for display. Colors distinguish curves within a panel and do not indicate matched identities across methods.}
    \label{fig:empiar10022-standard-predictions-small}
\end{figure*}

Figures~\ref{fig:empiar10022-standard-predictions-small}
and~\ref{fig:asyn-standard-predictions-small} show selected predictions,
with additional galleries in Appendix~\ref{sec:appendix-results}.

\begin{table*}[htb]
\centering
\caption{CryoTracer ablations of synthetic initialization, curve supervision, and architecture. Synthetic pretraining uses up to 20,000 steps; real-data training uses up to 5,000 steps unless marked 25k. Bold labels identify the main model configuration. Higher mF1 and Box IoU are better. Training and architecture details appear in Appendix~\ref{sec:experiment-series}.}
\label{tab:ablation-summary}
\small
\setlength{\tabcolsep}{4pt}
\renewcommand{\arraystretch}{1.08}
\resizebox{\textwidth}{!}{%
\begin{tabular}{@{}lllccrrrr@{}}
\toprule
 & & & & & \multicolumn{2}{c}{EMPIAR-10022} & \multicolumn{2}{c}{$\alpha$-synuclein} \\
\cmidrule(lr){6-7} \cmidrule(l){8-9}
Experiment & Training & Distance & Points & \shortstack{Pyramid\\levels} & mF1 & Box IoU & mF1 & Box IoU \\
\midrule
\multicolumn{9}{@{}l}{\textit{Standard regime}} \\
\addlinespace[2pt]
\textbf{Main} & \textbf{Synthetic + real} & \textbf{Fr\'echet} & \textbf{5} & \textbf{4} & 0.62 & 0.85 & 0.44 & 0.75 \\
\addlinespace
\multirow[c]{3}{*}{Training strategy} & Synthetic only & Fr\'echet & 5 & 4 & 0.00 & 0.05 & 0.00 & 0.01 \\
 & Real only (5k) & Fr\'echet & 5 & 4 & 0.16 & 0.49 & 0.18 & 0.51 \\
 & Real only (25k) & Fr\'echet & 5 & 4 & 0.59 & 0.83 & 0.39 & 0.71 \\
\addlinespace
\multirow[c]{2}{*}{Distance metric} & Real only (5k) & Chamfer & 5 & 4 & 0.00 & 0.42 & 0.00 & 0.35 \\
 & Synthetic + real & Chamfer & 5 & 4 & 0.03 & 0.71 & 0.01 & 0.69 \\
\addlinespace
\multirow[c]{2}{*}{Number of points} & Synthetic + real & Fr\'echet & 3 & 4 & 0.65 & 0.85 & 0.41 & 0.73 \\
 & Synthetic + real & Fr\'echet & 10 & 4 & 0.60 & 0.83 & 0.42 & 0.74 \\
\addlinespace
Feature pyramid & Synthetic + real & Fr\'echet & 5 & 1 & 0.56 & 0.81 & 0.36 & 0.73 \\
\addlinespace
Original attention & Synthetic + real & Fr\'echet & 5 & 4 & 0.57 & 0.82 & 0.33 & 0.69 \\
\addlinespace
\midrule
\multicolumn{9}{@{}l}{\textit{Practical regime}} \\
\addlinespace[2pt]
\textbf{Main} & \textbf{Synthetic + real} & \textbf{Fr\'echet} & \textbf{5} & \textbf{4} & 0.53 & 0.80 & 0.36 & 0.70 \\
\addlinespace
\multirow[c]{3}{*}{Training strategy} & Synthetic only & Fr\'echet & 5 & 4 & 0.00 & 0.05 & 0.00 & 0.01 \\
 & Real only (5k) & Fr\'echet & 5 & 4 & 0.07 & 0.45 & 0.02 & 0.28 \\
 & Real only (25k) & Fr\'echet & 5 & 4 & 0.40 & 0.68 & 0.11 & 0.44 \\
\addlinespace
\bottomrule
\end{tabular}%
}
\end{table*}

\begin{figure*}[t]
    \centering
    \includegraphics[width=\textwidth,height=0.82\textheight,keepaspectratio]{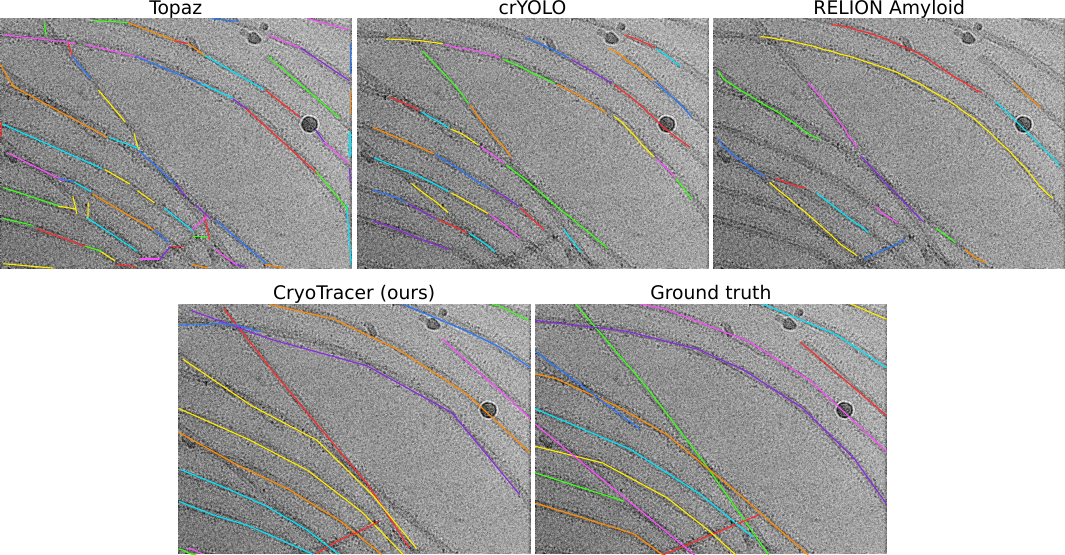}
    \caption{$\alpha$-synuclein, standard regime. One selected micrograph from the corresponding evaluation set, with method predictions and ground truth labeled in the panels. Images are low-pass filtered and contrast adjusted for display. Colors distinguish curves within a panel and do not indicate matched identities across methods.}
    \label{fig:asyn-standard-predictions-small}
\end{figure*}

\subsection{Ablations}
\label{sec:main-ablations}

In this section, we study how synthetic initialization, curve supervision,
and architecture affect filament detection. We compare
each variant with the main model in Table~\ref{tab:ablation-summary}. We
describe the training changes in Appendix~\ref{sec:experiment-series} and
report results at each matching tolerance in Appendix~\ref{sec:appendix-results}.

\paragraph{Synthetic initialization.}
With identical real-data partitions, 20,000 synthetic steps followed by
5,000 real-data steps yield standard-regime mF1 of $0.62$ on EMPIAR-10022
and $0.44$ on $\alpha$-synuclein. Training directly from RF-DETR for 25,000
real-data steps gives $0.59$ and $0.39$, respectively. The corresponding
practical-regime scores are $0.53$ and $0.36$ with synthetic initialization,
versus $0.40$ and $0.11$ for real-only training. The observed advantage is
therefore larger in the practical regime.

With only 5,000 real-only steps, mF1 is $0.16$ and $0.18$ in the standard
regime and $0.07$ and $0.02$ in the practical regime. Synthetic initialization
also improves spatial overlap relative to both real-only budgets.
Real-data adaptation remains necessary, as the frozen synthetic-only model
yields mF1 of $0.00$ in all four settings.

\paragraph{Curve supervision.}
Replacing Fr\'echet distance with Chamfer distance in both Hungarian matching
and regression reduces standard-regime mF1 from $0.62$ to
$0.03$ on EMPIAR-10022 and from $0.44$ to $0.01$ on $\alpha$-synuclein after
synthetic pretraining. Box IoU remains $0.71$ and $0.69$, respectively.
Substantial spatial overlap can therefore coexist with poor complete-instance
agreement. This is consistent with Chamfer's insensitivity to
point order.

\paragraph{Architecture.}
Restoring original RF-DETR box attention, while retaining five-point polyline
outputs, lowers standard-regime mF1 to
$0.57$ on EMPIAR-10022 and $0.33$ on $\alpha$-synuclein, compared with
$0.62$ and $0.44$ for the main model. Box IoU decreases to $0.82$ and
$0.69$, respectively.

Reducing the feature pyramid from four levels to one lowers standard-regime
mF1 from $0.62$ to $0.56$ on EMPIAR-10022 and from $0.44$ to $0.36$ on
$\alpha$-synuclein. Point-count effects vary by dataset. Three points score
highest on EMPIAR-10022, whereas five score highest on $\alpha$-synuclein.
The simpler three-point representation may suit EMPIAR-10022's relatively
clean, straight filaments. However, its two line segments limit the
representation of more strongly curved filaments, so this advantage may
not generalize to more complex geometries.
Ten points improve neither mF1 nor Box IoU over five.

We find that synthetic pretraining followed by real-data adaptation improves
mF1 over real-only training, with larger gains when annotations are limited.
Our results also support Fr\'echet supervision, the modified attention design,
and a four-level feature pyramid.
Although the preferred point count depends on the dataset, we find five
points to offer a reasonable compromise across the two datasets.

\subsection{Limitations}
\label{sec:limitations}

CryoTracer remains sensitive to ambiguous filament geometry. At crossings and overlaps, predicted traces can connect distinct filaments or recover only part of an instance. The fixed five-point representation may also limit the fidelity of traces with complex curvature. Synthetic pretraining reduces reliance on real training annotations but does not eliminate it. Our results cover two datasets with specimen-specific fine-tuning. Generalization to unseen specimens and acquisition conditions requires further evaluation. The next step is to test whether improved filament recovery translates into more effective downstream classification and reconstruction.

\section{Conclusion}

CryoTracer learns to trace individual cryo-EM filaments by predicting ordered
centerline polylines, making geometry and instance identity explicit in the
model output. The framework combines procedural synthetic pretraining,
orientation-invariant Fr\'echet supervision, and specimen-specific
fine-tuning. On EMPIAR-10022 and $\alpha$-synuclein, it achieves higher
complete-instance mF1 than Topaz and crYOLO in both standard and
limited-annotation regimes, and outperforms the RELION Amyloid
baseline on $\alpha$-synuclein. Ablations support the value of synthetic
initialization, with larger gains when real training annotations are scarce.
Together with the overlap metrics, these results show why filament tracing
should be evaluated for both spatial coverage and instance continuity.
We release new EMPIAR-10022 centerline annotations
to support further evaluation.

\clearpage







\subsubsection*{Acknowledgments}
Computations were performed at the \href{https://www.wcss.pl}{Wrocław Centre for Networking and Supercomputing} under grant No.~1766404231.

\bibliography{bibliography}
\bibliographystyle{iclr2027_conference}

\appendix

\clearpage

\begin{center}
    \LARGE{
CryoTracer}\\
\Large{Appendix}
\end{center}

\vspace{30px}

\section{Data}
\label{sec:appendix-data}

We use two real cryo-electron microscopy datasets and a procedurally generated
dataset. All annotations are represented as ordered polylines, where each
polyline describes the centerline of one filament.

\paragraph{EMPIAR-10022.}
EMPIAR-10022 is a public dataset of tobacco mosaic virus
micrographs~\cite{FROMM201587}. The original helical annotations were of
insufficient quality for reliable filament-level evaluation, so we
reannotated the micrographs with ordered filament centerlines. We make these
new annotations publicly available. Our annotated dataset contains 109 images at
a native resolution of $4096 \times 4096$ pixels and a pixel size of
$1.06\,\text{\AA}/\text{px}$.

\paragraph{$\alpha$-synuclein.}
The $\alpha$-synuclein dataset is a private, in-house collection of 11
reconstruction sessions comprising 325 cryo-EM micrographs annotated by a
cryo-EM expert. All sessions were collected by the same person in the same
laboratory. Each micrograph has a native resolution of
$5760 \times 4092$ pixels and a pixel size of $0.65\,\text{\AA}/\text{px}$.

\paragraph{Real-image preprocessing.}
Micrographs are downsampled by a factor of 4 using bicubic interpolation with
antialiasing (\texttt{torchvision} resize with
\texttt{antialias=True}). EMPIAR-10022 images are mapped from
$4096 \times 4096$ to $1024 \times 1024$, and $\alpha$-synuclein images from
$5760 \times 4092$ to $1440 \times 1023$. Polyline coordinates are divided by
4 and normalized by the resulting image dimensions. Pixel-size metadata is
multiplied by 4. Training applies horizontal and vertical flips independently with
probability $0.5$ and uniformly sampled in-plane rotations. A radial Fourier
low-pass mask at $0.2$\,\AA$^{-1}$ filters the images, which are standardized
over valid pixels and clipped to $[-5,5]$. Validation and test preprocessing
uses the same filtering and normalization without geometric augmentation.

\paragraph{Synthetic pretraining data.}
To expose the model to a broad range of filament geometries before adaptation
to real data, we construct a procedural micrograph dataset. Each example is
defined in a native $4096 \times 4096$ coordinate system and rendered directly
at $1024 \times 1024$ resolution. Non-empty images contain between one and 30
curved filaments. Filaments are generated from smooth parametric curves, clipped
at the image boundary, and include explicit crossing configurations. Ten
percent of examples are empty, allowing the model to learn from background-only
micrographs.

We randomize both imaging and specimen properties. The native pixel size is
sampled uniformly from $0.65$ to $1.5\,\text{\AA}/\text{px}$, corresponding to
an effective input sampling of $2.6$-$6.0\,\text{\AA}/\text{px}$ at the
rendered resolution. Filament widths are sampled from
$100$-$200\,\text{\AA}$. Appearance variation includes both contrast
polarities, spatial background gradients, Gaussian blur, and additive noise.
These perturbations are intended to encourage transfer across datasets rather
than reproduce a particular microscope contrast-transfer function.

\subsection{Polyline sampling}
\label{sec:polyline-sampling}

Synthetic and real annotations use the same five-point target conversion.
For annotations with more than five points, we select five vertices at
uniformly spaced indices, rounded down, including both endpoints.
For annotations with at most five points, we retain every annotated point
and add any needed points along the
existing segments in proportion to their lengths. Thus, short annotations are
only augmented rather than resampled, so no annotated bend is inadvertently
omitted or replaced by a chord.

For evaluation, predicted and reference polylines are sampled uniformly by
arc length to 32 points. This gives curves with different original point
counts a common representation for matching and metric computation while
preserving a denser approximation of local bends.

\begin{figure*}[t]
    \centering
    \includegraphics[width=0.9\textwidth,height=0.82\textheight,keepaspectratio]{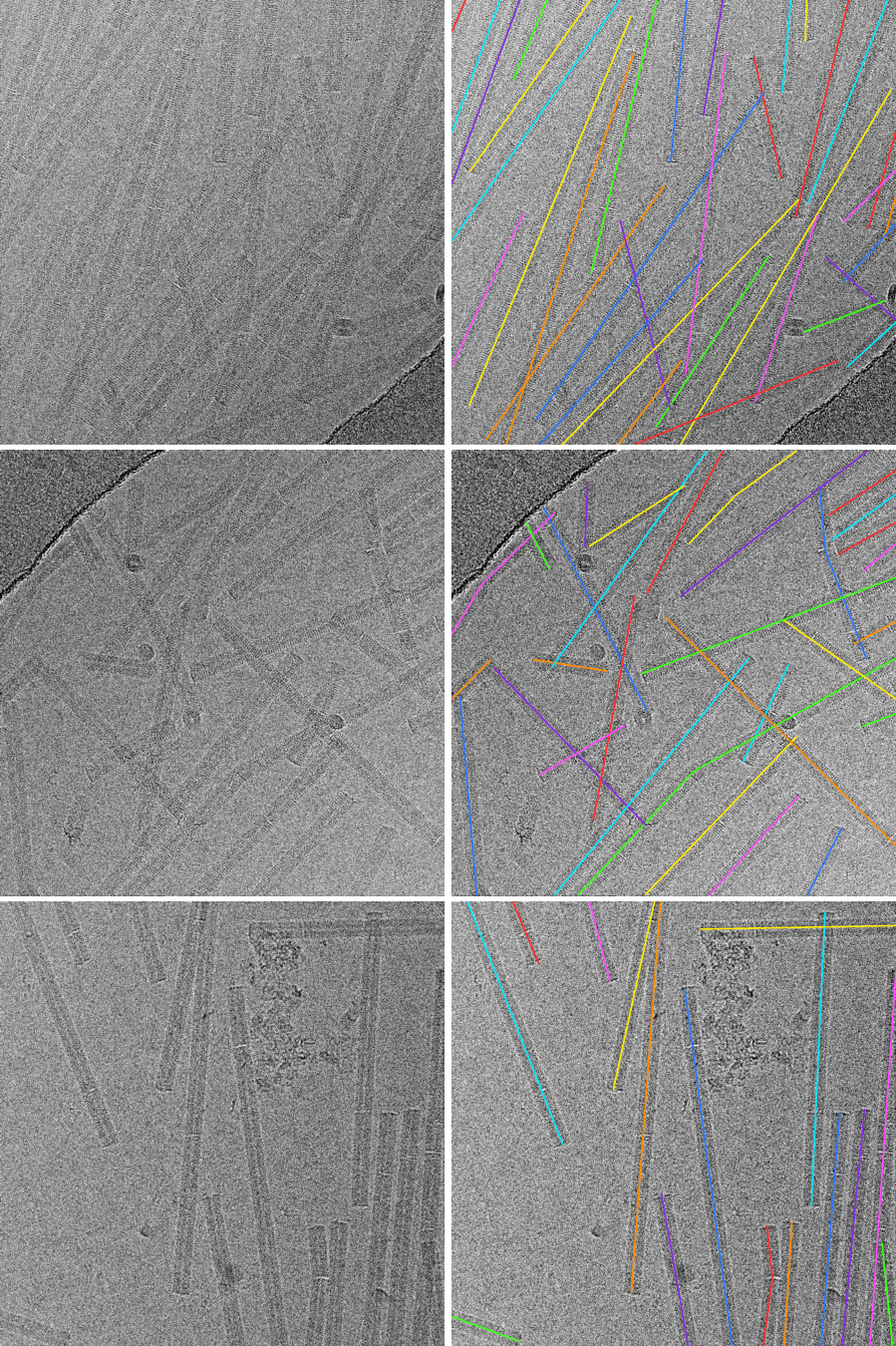}
    \caption{EMPIAR-10022 dataset examples. Each row pairs an image (left) with a low-pass-filtered image and its ground-truth centerlines (right). Colors distinguish annotated filament instances.}
    \label{fig:empiar10022-samples}
\end{figure*}

\begin{figure*}[t]
    \centering
    \includegraphics[width=0.9\textwidth,height=0.82\textheight,keepaspectratio]{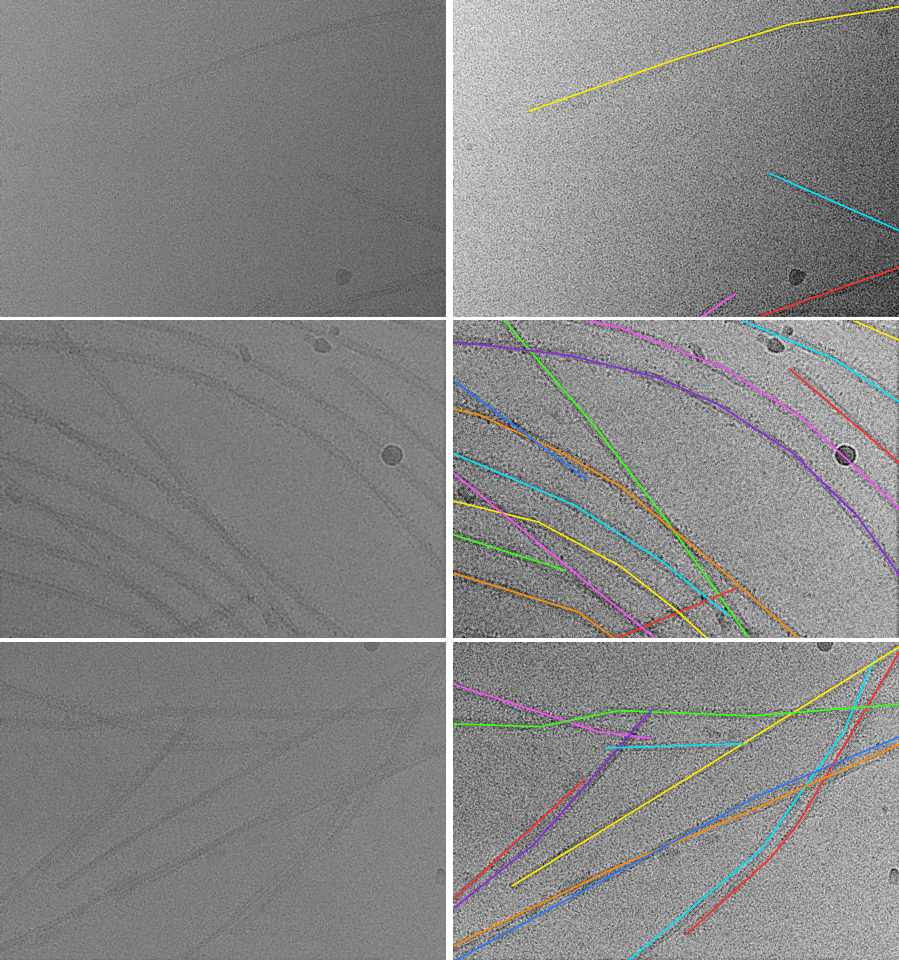}
    \caption{$\alpha$-synuclein dataset examples. Each row pairs an image (left) with a low-pass-filtered image and its ground-truth centerlines (right). Colors distinguish annotated filament instances.}
    \label{fig:asyn-samples}
\end{figure*}

\begin{figure*}[t]
    \centering
    \includegraphics[width=0.9\textwidth,height=0.82\textheight,keepaspectratio]{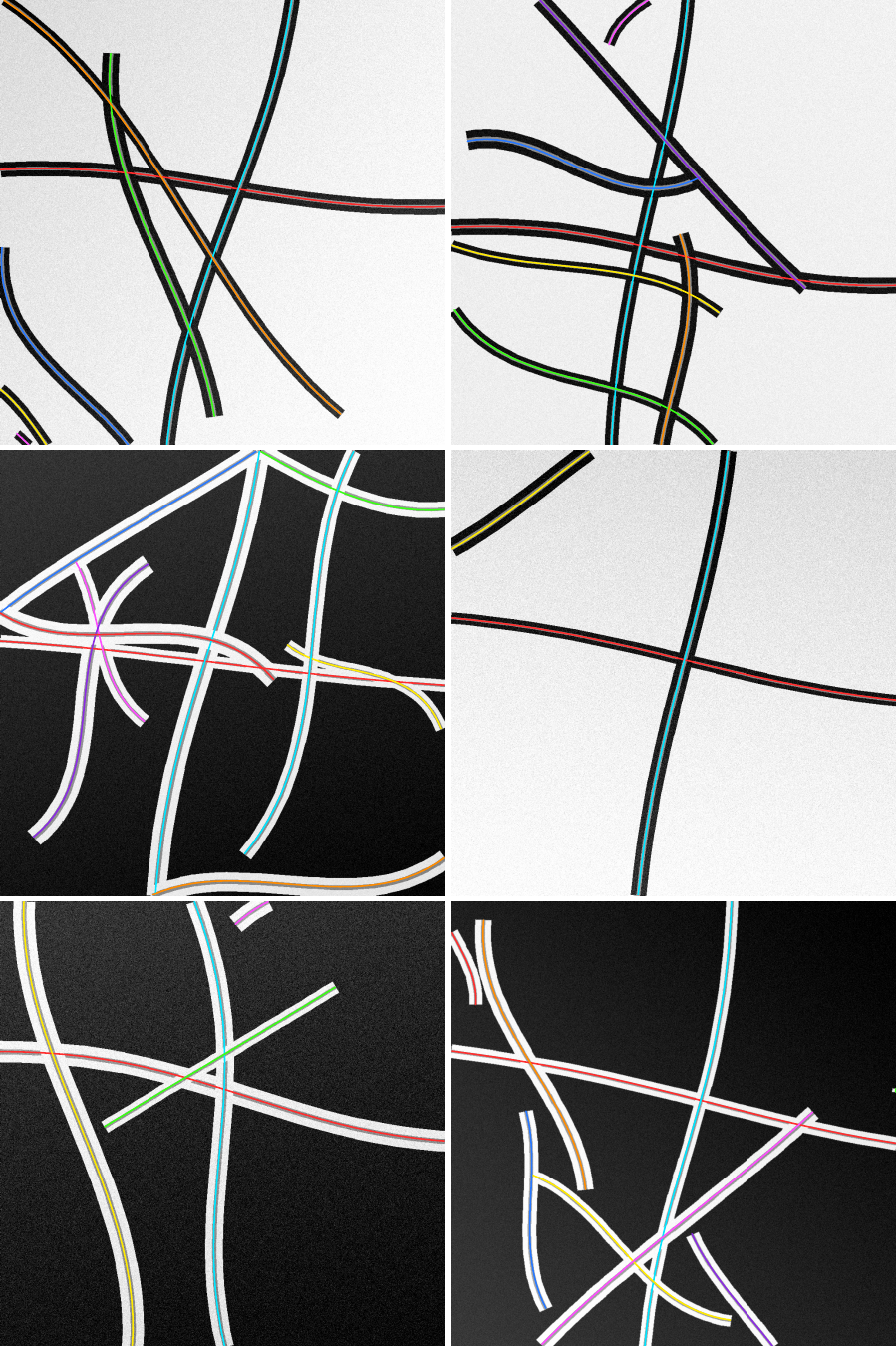}
    \caption{Synthetic dataset examples. Six synthetic micrographs are shown with their ground-truth centerlines. Colors distinguish annotated filament instances.}
    \label{fig:synthetic-samples}
\end{figure*}

\clearpage
\section{Training and evaluation details}
\label{sec:implementation-details}

\subsection{Training configuration}

\paragraph{Hardware and optimization.}
Each CryoTracer run uses one NVIDIA H100 GPU, bfloat16 mixed precision,
gradient checkpointing, and a batch size of 64 micrographs. AdamW uses weight
decay $10^{-4}$ and initial learning rates of $10^{-4}$ for the detector and
$10^{-5}$ for the backbone. A cosine schedule decays both towards $10^{-6}$
over each stage, with gradient clipping at $1.0$. The complete model remains
trainable throughout.

\paragraph{Training stages and validation.}
Synthetic pretraining starts from RF-DETR weights adapted to single-channel
input and runs for at most 20,000 steps, with validation every 500 steps.
The selected synthetic model initializes dataset-specific fine-tuning for at
most 5,000 steps, with validation every 100 steps on the corresponding real
validation partition. Fine-tuning starts a new optimizer and learning-rate
schedule. Within each seed, the same selected synthetic initialization is
used for both real datasets and their standard and practical regimes.

\paragraph{Checkpoint selection.}
At each stage, we retain the checkpoint minimizing the final decoder's
classification loss plus weighted polyline regression loss on the validation
set. Encoder and intermediate decoder terms are excluded from this selection
criterion.
Final evaluation uses the selected real-data checkpoint. Test metrics do not
influence model selection.

\paragraph{Loss implementation.}
\label{sec:loss-implementation}
Batch regression losses are normalized by the total target count, clamped to
at least one. Distributed training synchronizes this normalization across
workers. Squared distances are floored at $10^{-12}$ before taking square
roots. The hard minimum and maximum operations in the Fr\'echet recurrence
are piecewise differentiable. Automatic differentiation follows the selected
branches with the framework's subgradient convention at ties. Reversal
invariance applies to the objective, not to the decoder's internal features.

\paragraph{Ablation configurations.}
\label{sec:experiment-series}
The ablation reference uses the main model's configuration, including
synthetic pretraining, Fr\'echet matching and regression, auxiliary decoder
supervision, five polyline points, and a four-level feature pyramid. The no-synthetic
condition starts directly from RF-DETR weights with a 5,000-step real-data
training budget. Additional real-only controls train for 25,000 steps with
the cosine schedule extended over that budget, matching the total optimizer
steps of 20,000-step synthetic pretraining plus 5,000-step fine-tuning. These
controls retain the same model, data partitions, and validation-selection
criterion in both regimes. Optimizer-step matching does not imply identical
wall time or computational cost. The Chamfer condition changes both matching
and regression in pretraining and fine-tuning, using its own
validation-selected synthetic initialization. The architecture ablations use
three or ten polyline points, or a single feature-pyramid level. Each is
pretrained and fine-tuned with the corresponding architecture.

The original-attention ablation restores RF-DETR's box-based deformable
cross-attention and box positional encoding. At each decoder layer, the
reference polyline supplies its axis-aligned enclosing box. This variant
samples four locations per head and feature level, compared with four per
polyline point (20 total) in the main model. It retains five-point outputs,
polyline refinement and losses, and the four-level feature pyramid, using
its own 20,000-step synthetic pretraining and 5,000-step real-data fine-tuning.
The comparison therefore changes reference geometry, positional encoding,
and sampling budget together.

\subsection{Inference and particle export}
\label{sec:inference-export}

\paragraph{Duplicate suppression.}
For filament evaluation, polyline NMS identifies duplicates using their
unoriented directions, projected overlap relative to the shorter projected
trace, and perpendicular separation. We use a $12^\circ$ angular threshold,
minimum overlap ratio $0.8$, and maximum separation $140$\,\AA{} across both
datasets and regimes. Suppression favors the higher-confidence curve and
preserves its geometry. Fragments remain separate.

\paragraph{RELION particle export.}
The optional export samples confidence-filtered curves uniformly by arc
length, with physical spacing equal to the user-specified helical rise times
the number of asymmetric units between samples. Centers whose extraction
boxes would extend outside the micrograph are omitted. Each particle records
its native coordinates, parent filament identifier, local tangent-angle prior,
and cumulative distance along the original trace. Export-specific duplicate
suppression and optional crossing exclusion operate on sampled particles.
Removed samples leave gaps without resetting track distances. Filament
identifiers restart within each micrograph. The resulting STAR file includes
the associated optics and CTF metadata from the supplied micrographs STAR
file and supports subsequent particle extraction in RELION. CryoTracer writes
particle metadata, and RELION performs image extraction.

\subsection{Evaluation protocol}
\label{sec:experimental-protocol}

\paragraph{Repeated runs and aggregation.}
Main-text CryoTracer results report means over five training runs (seeds
42-46). Appendix tables add the sample standard deviation. Ablations use
seed 42, and each baseline has one run. Within each seed, instance scores
are averaged over the same evaluation micrographs within each regime.
mF1, mRecall, and mPrecision additionally average over distance tolerances
before aggregation across runs. Pick counts and mean pick lengths are also
computed separately for each seed before reporting their mean and sample
standard deviation. Standard deviations describe variation across seeds,
not across tolerances, micrographs, or individual filament lengths.

\paragraph{Data regimes.}
We evaluate two data regimes. In the standard regime, EMPIAR-10022 is split into 76 training, 16 validation, and 17 test micrographs, while the $\alpha$-synuclein dataset is split into 227 training, 48 validation, and 50 test micrographs. These correspond approximately to a 70/15/15\% train/validation/test split.

The practical regime reduces the amount of annotated data available for training to approximately 15\% of each dataset while retaining approximately 15\% for validation and leaving about 70\% for evaluation. We obtain this split by rotating the standard partitions. The validation set is used for training, the test set for validation, and the training set for evaluation. This results in $16/17/76$ train/validation/evaluation micrographs for EMPIAR-10022 and $48/50/227$ for $\alpha$-synuclein.

Partitions are defined at the micrograph level. For $\alpha$-synuclein,
reconstruction sessions overlap across training, validation, and evaluation
partitions in both regimes, while individual micrographs remain disjoint
within each regime. This protocol evaluates held-out micrographs from the
collection described in Appendix~\ref{sec:appendix-data}.

\paragraph{Comparison methods.}
\label{sec:reference-methods}
Topaz and crYOLO use the same supervised partitions as CryoTracer.
Their configuration required substantial manual tuning by a cryo-EM expert,
including dataset-specific picking and filament-tracing settings, which were
then fixed for evaluation. Topaz is trained for 10 epochs with an expected particle count
of 200, using downsampling factors of four and six and extraction thresholds
of $-3$ and $-4$ for EMPIAR-10022 and $\alpha$-synuclein, respectively. crYOLO uses
filament mode with expert-selected box sizes of 400 and 512 pixels,
approximately twice the filament width to include surrounding image context.
The filament-width argument (\texttt{-fw}) is set to the corresponding box
size, with tracing box distances of 40 and 51 pixels, respectively. These
expert-selected picking and tracing settings are independent of the tighter
physical filament widths used for spatial-overlap evaluation below.
CryoTracer uses the same optimization recipe and inference thresholds across
both datasets and regimes, without corresponding dataset-specific manual
tuning. Its specimen-specific model-weight fine-tuning remains part of
training. This comparison of tuning effort is qualitative. The experiments
compare configured systems without matched hyperparameter-search or compute
budgets, and do not quantify expert time or the number of tuning trials.

\paragraph{RELION Amyloid.}
RELION 5.1's AmyTracer pipeline combines an amyloid-specific Fourier signal
with a pretrained neural network and subsequent tracing
\cite{lovestam2026relion}. It is therefore a learned, pretrained reference,
even when used without dataset-specific training.
We evaluate the official pretrained AmyTracer v2.0 checkpoint only on
$\alpha$-synuclein.
The RELION-5.1.0 release supplies inference code but no AmyTracer training
or fine-tuning code, so we perform no dataset-specific weight updates.
The same checkpoint is evaluated on the original test partition in the
standard regime and the original training partition in the practical regime,
matching the evaluation micrographs used by the other methods within each
regime. We also retain the RELION-5.1 default tracing parameters.
Its amyloid cross-beta signal makes it applicable to $\alpha$-synuclein,
but not the tobacco mosaic virus filaments in EMPIAR-10022.

\paragraph{Inference settings and representations.}
CryoTracer uses a confidence threshold of $0.5$ and the polyline NMS described in
Appendix~\ref{sec:inference-export}. Topaz predictions are interpreted as
consecutive endpoint pairs. crYOLO box centers and RELION traces are grouped
by filament identifier.
These conversions preserve the methods' exported representations. Short
fragments can therefore achieve high spatial overlap while failing to match
a complete annotated instance. All coordinates are mapped to native pixels
and converted to physical units for evaluation.

\subsection{Evaluation metric definitions}
\label{sec:evaluation-metric-details}

\paragraph{Instance matching.}
Predicted and reference curves are sampled uniformly by arc length to 32
points. The distinction from the five-point training construction is explained
in Appendix~\ref{sec:polyline-sampling}. Following the notation in
Section~\ref{sec:method}, let $\mathcal P=\{\hat{\mathbf P}_i\}$ and
$\mathcal G=\{\mathbf Y_j\}$ denote the predicted and ground truth polylines,
respectively, and define
\begin{equation}
    D_{ij}=\bar{d}_F(\hat{\mathbf P}_i,\mathbf Y_j),
\end{equation}
using the orientation-invariant distance in
Equation~\ref{eq:oriented-frechet}, measured in \AA{}. At tolerance $\tau$,
only pairs with $D_{ij}\leq\tau$ are eligible. Hungarian assignment with this
tolerance constraint first maximizes the number of valid one-to-one matches,
then minimizes their total distance. Matched pairs are true positives.
Unmatched predictions and ground truth polylines are false positives and
false negatives, respectively. The per-image counts give
\begin{equation}
    \begin{aligned}
        P_\tau &= \frac{\mathrm{TP}_\tau}
                           {\mathrm{TP}_\tau+\mathrm{FP}_\tau},\\
        R_\tau &= \frac{\mathrm{TP}_\tau}
                           {\mathrm{TP}_\tau+\mathrm{FN}_\tau},\\
        F1_\tau &= \frac{2P_\tau R_\tau}{P_\tau+R_\tau}.
    \end{aligned}
\end{equation}
We report the mean of each per-image score, with undefined ratios assigned
zero, including images with no predictions or annotations. Macro F1 is
therefore the mean of per-image F1 scores. With the fixed tolerance set
$\mathcal T=\{25,50,100,150,200\}$\,\AA{}, our primary metric is
\begin{equation}
    \operatorname{mF1}
    = \frac{1}{|\mathcal T|}\sum_{\tau\in\mathcal T}
      \frac{1}{M}\sum_{i=1}^{M} F1_{i,\tau},
\end{equation}
where $M$ is the number of evaluation micrographs. mRecall and mPrecision
use the same equal-weight averaging of the corresponding per-image scores.
mF1 averages F1 directly and is not the harmonic mean of mPrecision and
mRecall. Scores at individual tolerances are secondary diagnostics.

\paragraph{Spatial overlap.}
For each image, let $B_{\mathcal P}$ and $B_{\mathcal G}$ be the unions of
axis-aligned boxes swept along the predicted and reference curves. The metric
box width is independent of the particle-extraction box. We use 170 native
pixels for EMPIAR-10022, corresponding to the established $180$\,\AA{} TMV
diameter, and 215 pixels for $\alpha$-synuclein, corresponding to an assumed
$140$\,\AA{} filament width. These tight widths measure agreement around the
filament itself and are shared by all methods within each dataset. Define
\begin{equation}
    \begin{aligned}
        \operatorname{BoxIoU}
        &= \frac{|B_{\mathcal P}\cap B_{\mathcal G}|}
                 {|B_{\mathcal P}\cup B_{\mathcal G}|},\\
        \operatorname{Coverage}
        &= \frac{|B_{\mathcal P}\cap B_{\mathcal G}|}
                 {|B_{\mathcal G}|}.
    \end{aligned}
\end{equation}
Box IoU penalizes both missing and excess predicted area. Coverage measures
the fraction of reference area recovered. We average Box IoU over all images,
assigning one when both masks are empty, and coverage over images with
non-empty ground truth. These union-based metrics do not measure instance
identity and depend on the specified box width.

\clearpage
\section{Additional results}
\label{sec:appendix-results}

\begin{table*}[!htbp]
\centering
\caption{Detailed filament-detection results on EMPIAR-10022. Precision (P), recall (R), and F1 are averaged over micrographs at each distance tolerance (in \AA{}). Box IoU and coverage measure spatial overlap. Pick count and mean polyline length (nm) summarize filament traces, with ground truth as a reference. CryoTracer values are mean $\pm$ sample standard deviation over five seeds, with statistics computed separately per seed. Bold marks the highest unrounded score per metric and regime. Pick statistics are descriptive. Regimes use different evaluation sets.}
\label{tab:empiar10022-full}
\resizebox{\textwidth}{!}{%
\begin{tabular}{ll|rrr|rrr|rrr|rrr|rrr|rr|rr}
\toprule
 &  & P@25 & R@25 & F1@25 & P@50 & R@50 & F1@50 & P@100 & R@100 & F1@100 & P@150 & R@150 & F1@150 & P@200 & R@200 & F1@200 & Box IoU & Coverage & Pick count & Mean pick length (nm) \\
Regime & Method &  &  &  &  &  &  &  &  &  &  &  &  &  &  &  &  &  &  &  \\
\midrule
\multirow[c]{4}{*}{standard} & CryoTracer (ours) & \textbf{0.25 $\pm$ 0.04} & \textbf{0.25 $\pm$ 0.04} & \textbf{0.25 $\pm$ 0.04} & \textbf{0.55 $\pm$ 0.02} & \textbf{0.55 $\pm$ 0.02} & \textbf{0.55 $\pm$ 0.02} & \textbf{0.74 $\pm$ 0.02} & \textbf{0.74 $\pm$ 0.02} & \textbf{0.74 $\pm$ 0.02} & \textbf{0.80 $\pm$ 0.02} & \textbf{0.79 $\pm$ 0.02} & \textbf{0.79 $\pm$ 0.02} & \textbf{0.82 $\pm$ 0.02} & \textbf{0.82 $\pm$ 0.01} & \textbf{0.82 $\pm$ 0.02} & \textbf{0.84 $\pm$ 0.01} & \textbf{0.90 $\pm$ 0.01} & 337 $\pm$ 4 & 176 $\pm$ 3 \\
 & Topaz & 0.01 & 0.02 & 0.01 & 0.02 & 0.04 & 0.03 & 0.06 & 0.13 & 0.08 & 0.09 & 0.22 & 0.13 & 0.11 & 0.25 & 0.15 & 0.82 & 0.88 & 947 & 56 \\
 & crYOLO & 0.01 & 0.01 & 0.01 & 0.01 & 0.01 & 0.01 & 0.02 & 0.02 & 0.02 & 0.04 & 0.03 & 0.04 & 0.06 & 0.05 & 0.06 & 0.66 & 0.69 & 329 & 122 \\
 & Ground truth & \textemdash{} & \textemdash{} & \textemdash{} & \textemdash{} & \textemdash{} & \textemdash{} & \textemdash{} & \textemdash{} & \textemdash{} & \textemdash{} & \textemdash{} & \textemdash{} & \textemdash{} & \textemdash{} & \textemdash{} & \textemdash{} & \textemdash{} & 347 & 181 \\
\cline{1-21}
\multirow[c]{4}{*}{practical} & CryoTracer (ours) & \textbf{0.14 $\pm$ 0.04} & \textbf{0.13 $\pm$ 0.04} & \textbf{0.14 $\pm$ 0.04} & \textbf{0.45 $\pm$ 0.06} & \textbf{0.43 $\pm$ 0.05} & \textbf{0.44 $\pm$ 0.05} & \textbf{0.67 $\pm$ 0.05} & \textbf{0.64 $\pm$ 0.04} & \textbf{0.65 $\pm$ 0.04} & \textbf{0.74 $\pm$ 0.04} & \textbf{0.71 $\pm$ 0.03} & \textbf{0.72 $\pm$ 0.04} & \textbf{0.78 $\pm$ 0.04} & \textbf{0.75 $\pm$ 0.03} & \textbf{0.76 $\pm$ 0.03} & 0.81 $\pm$ 0.01 & 0.86 $\pm$ 0.01 & 1,474 $\pm$ 34 & 184 $\pm$ 5 \\
 & Topaz & 0.01 & 0.04 & 0.02 & 0.03 & 0.09 & 0.05 & 0.06 & 0.17 & 0.09 & 0.09 & 0.24 & 0.13 & 0.10 & 0.28 & 0.15 & \textbf{0.86} & \textbf{0.92} & 4,607 & 61 \\
 & crYOLO & 0.00 & 0.00 & 0.00 & 0.00 & 0.00 & 0.00 & 0.02 & 0.02 & 0.02 & 0.03 & 0.03 & 0.03 & 0.04 & 0.03 & 0.04 & 0.65 & 0.68 & 1,559 & 118 \\
 & Ground truth & \textemdash{} & \textemdash{} & \textemdash{} & \textemdash{} & \textemdash{} & \textemdash{} & \textemdash{} & \textemdash{} & \textemdash{} & \textemdash{} & \textemdash{} & \textemdash{} & \textemdash{} & \textemdash{} & \textemdash{} & \textemdash{} & \textemdash{} & 1,578 & 190 \\
\cline{1-21}
\bottomrule
\end{tabular}%
}
\end{table*}
\begin{table*}[p]
\centering
\caption{Detailed filament-detection results on $\alpha$-synuclein. Precision (P), recall (R), and F1 are averaged over micrographs at each distance tolerance (in \AA{}). Box IoU and coverage measure spatial overlap. Pick count and mean polyline length (nm) summarize filament traces, with ground truth as a reference. CryoTracer values are mean $\pm$ sample standard deviation over five seeds, with statistics computed separately per seed. Bold marks the highest unrounded score per metric and regime. Pick statistics are descriptive. Regimes use different evaluation sets.}
\label{tab:asyn-full}
\resizebox{\textwidth}{!}{%
\begin{tabular}{ll|rrr|rrr|rrr|rrr|rrr|rr|rr}
\toprule
 &  & P@25 & R@25 & F1@25 & P@50 & R@50 & F1@50 & P@100 & R@100 & F1@100 & P@150 & R@150 & F1@150 & P@200 & R@200 & F1@200 & Box IoU & Coverage & Pick count & Mean pick length (nm) \\
Regime & Method &  &  &  &  &  &  &  &  &  &  &  &  &  &  &  &  &  &  &  \\
\midrule
\multirow[c]{5}{*}{standard} & CryoTracer (ours) & \textbf{0.02 $\pm$ 0.01} & \textbf{0.03 $\pm$ 0.02} & \textbf{0.03 $\pm$ 0.02} & \textbf{0.18 $\pm$ 0.04} & \textbf{0.20 $\pm$ 0.04} & \textbf{0.19 $\pm$ 0.04} & \textbf{0.47 $\pm$ 0.05} & \textbf{0.51 $\pm$ 0.06} & \textbf{0.48 $\pm$ 0.05} & \textbf{0.62 $\pm$ 0.04} & \textbf{0.67 $\pm$ 0.04} & \textbf{0.63 $\pm$ 0.04} & \textbf{0.68 $\pm$ 0.03} & \textbf{0.73 $\pm$ 0.01} & \textbf{0.69 $\pm$ 0.02} & \textbf{0.75 $\pm$ 0.01} & \textbf{0.84 $\pm$ 0.01} & 278 $\pm$ 11 & 176 $\pm$ 7 \\
 & Topaz & 0.00 & 0.00 & 0.00 & 0.00 & 0.01 & 0.00 & 0.01 & 0.05 & 0.01 & 0.01 & 0.07 & 0.02 & 0.01 & 0.08 & 0.02 & 0.59 & 0.80 & 1,659 & 33 \\
 & crYOLO & 0.00 & 0.00 & 0.00 & 0.01 & 0.01 & 0.01 & 0.01 & 0.01 & 0.01 & 0.01 & 0.01 & 0.01 & 0.02 & 0.02 & 0.02 & 0.60 & 0.66 & 540 & 60 \\
 & RELION Amyloid & 0.00 & 0.00 & 0.00 & 0.00 & 0.00 & 0.00 & 0.13 & 0.12 & 0.12 & 0.20 & 0.20 & 0.20 & 0.23 & 0.23 & 0.22 & 0.59 & 0.63 & 342 & 94 \\
 & Ground truth & \textemdash{} & \textemdash{} & \textemdash{} & \textemdash{} & \textemdash{} & \textemdash{} & \textemdash{} & \textemdash{} & \textemdash{} & \textemdash{} & \textemdash{} & \textemdash{} & \textemdash{} & \textemdash{} & \textemdash{} & \textemdash{} & \textemdash{} & 272 & 203 \\
\cline{1-21}
\multirow[c]{5}{*}{practical} & CryoTracer (ours) & \textbf{0.01 $\pm$ 0.00} & \textbf{0.01 $\pm$ 0.00} & \textbf{0.01 $\pm$ 0.00} & \textbf{0.12 $\pm$ 0.03} & \textbf{0.12 $\pm$ 0.03} & \textbf{0.12 $\pm$ 0.03} & \textbf{0.40 $\pm$ 0.05} & \textbf{0.43 $\pm$ 0.03} & \textbf{0.40 $\pm$ 0.04} & \textbf{0.54 $\pm$ 0.04} & \textbf{0.58 $\pm$ 0.01} & \textbf{0.54 $\pm$ 0.03} & \textbf{0.61 $\pm$ 0.05} & \textbf{0.66 $\pm$ 0.01} & \textbf{0.62 $\pm$ 0.03} & \textbf{0.69 $\pm$ 0.01} & 0.80 $\pm$ 0.01 & 1,374 $\pm$ 97 & 181 $\pm$ 9 \\
 & Topaz & 0.00 & 0.00 & 0.00 & 0.00 & 0.01 & 0.00 & 0.01 & 0.05 & 0.02 & 0.02 & 0.07 & 0.03 & 0.02 & 0.11 & 0.04 & 0.63 & \textbf{0.83} & 7,891 & 34 \\
 & crYOLO & 0.00 & 0.00 & 0.00 & 0.00 & 0.00 & 0.00 & 0.00 & 0.00 & 0.00 & 0.00 & 0.00 & 0.00 & 0.00 & 0.01 & 0.01 & 0.58 & 0.62 & 2,498 & 61 \\
 & RELION Amyloid & 0.00 & 0.00 & 0.00 & 0.00 & 0.00 & 0.00 & 0.07 & 0.08 & 0.07 & 0.16 & 0.18 & 0.16 & 0.19 & 0.21 & 0.20 & 0.61 & 0.66 & 1,914 & 88 \\
 & Ground truth & \textemdash{} & \textemdash{} & \textemdash{} & \textemdash{} & \textemdash{} & \textemdash{} & \textemdash{} & \textemdash{} & \textemdash{} & \textemdash{} & \textemdash{} & \textemdash{} & \textemdash{} & \textemdash{} & \textemdash{} & \textemdash{} & \textemdash{} & 1,364 & 203 \\
\cline{1-21}
\bottomrule
\end{tabular}%
}
\end{table*}
\begin{table*}[p]
\centering
\setlength{\tabcolsep}{3pt}
\caption{CryoTracer ablations on EMPIAR-10022. Precision (P), recall (R), and F1 are averaged over micrographs at each distance tolerance (in \AA{}). Box IoU and coverage measure spatial overlap. Synthetic pretraining uses up to 20,000 steps. Real-data training uses up to 5,000 steps unless marked 25k. Synthetic-only rows evaluate the same frozen checkpoint in both regimes. Bold labels identify the main model configuration. Bold scores mark the highest unrounded value per metric and regime. Higher scores are better. Training and architecture details appear in Appendix~\ref{sec:experiment-series}.}
\label{tab:empiar10022-ablation}
\resizebox{\textwidth}{!}{%
\begin{tabular}{lllllrrrrrrrrrrrrrrrrr}
\toprule
 &  &  &  &  & P@25 & R@25 & F1@25 & P@50 & R@50 & F1@50 & P@100 & R@100 & F1@100 & P@150 & R@150 & F1@150 & P@200 & R@200 & F1@200 & Box IoU & Coverage \\
Regime & Experiment & training & distance & architecture &  &  &  &  &  &  &  &  &  &  &  &  &  &  &  &  &  \\
\midrule
\multirow[c]{10}{*}{standard} & \textbf{Main} & \textbf{Synthetic + real} & \textbf{Fréchet} & \textbf{5 points, 4 levels} & 0.23 & 0.23 & 0.23 & 0.54 & 0.54 & 0.54 & 0.74 & 0.73 & 0.73 & 0.79 & 0.79 & 0.79 & 0.83 & 0.82 & 0.82 & 0.85 & 0.90 \\
\cline{2-22} \cline{3-22} \cline{4-22}
 & \multirow[c]{3}{*}{Training strategy} & Synthetic only & Fréchet & 5 points, 4 levels & 0.00 & 0.00 & 0.00 & 0.00 & 0.00 & 0.00 & 0.00 & 0.00 & 0.00 & 0.00 & 0.00 & 0.00 & 0.00 & 0.00 & 0.00 & 0.05 & 0.05 \\
\cline{3-22} \cline{4-22}
 &  & Real only & Fréchet & 5 points, 4 levels & 0.00 & 0.00 & 0.00 & 0.03 & 0.02 & 0.02 & 0.16 & 0.12 & 0.13 & 0.33 & 0.23 & 0.27 & 0.45 & 0.31 & 0.37 & 0.49 & 0.56 \\
\cline{3-22} \cline{4-22}
 &  & Real only (25k) & Fréchet & 5 points, 4 levels & 0.25 & 0.24 & 0.24 & 0.51 & 0.50 & 0.50 & 0.70 & 0.68 & 0.69 & 0.77 & 0.74 & 0.75 & 0.80 & 0.78 & 0.79 & 0.83 & 0.88 \\
\cline{2-22} \cline{3-22} \cline{4-22}
 & \multirow[c]{2}{*}{Distance metric} & Real only & Chamfer & 5 points, 4 levels & 0.00 & 0.00 & 0.00 & 0.00 & 0.00 & 0.00 & 0.00 & 0.00 & 0.00 & 0.00 & 0.00 & 0.00 & 0.00 & 0.00 & 0.00 & 0.42 & 0.60 \\
\cline{3-22} \cline{4-22}
 &  & Synthetic + real & Chamfer & 5 points, 4 levels & 0.00 & 0.00 & 0.00 & 0.00 & 0.00 & 0.00 & 0.02 & 0.02 & 0.02 & 0.05 & 0.05 & 0.05 & 0.08 & 0.08 & 0.08 & 0.71 & 0.81 \\
\cline{2-22} \cline{3-22} \cline{4-22}
 & \multirow[c]{2}{*}{Number of points} & \multirow[c]{2}{*}{Synthetic + real} & \multirow[c]{2}{*}{Fréchet} & 3 points, 4 levels & \textbf{0.28} & \textbf{0.28} & \textbf{0.28} & \textbf{0.57} & \textbf{0.57} & \textbf{0.57} & \textbf{0.76} & \textbf{0.76} & \textbf{0.76} & \textbf{0.81} & \textbf{0.81} & \textbf{0.81} & \textbf{0.83} & \textbf{0.83} & \textbf{0.83} & \textbf{0.85} & \textbf{0.90} \\
 &  &  &  & 10 points, 4 levels & 0.18 & 0.18 & 0.18 & 0.50 & 0.50 & 0.49 & 0.74 & 0.74 & 0.74 & 0.79 & 0.79 & 0.79 & 0.82 & 0.82 & 0.82 & 0.83 & 0.88 \\
\cline{2-22} \cline{3-22} \cline{4-22}
 & Feature pyramid & Synthetic + real & Fréchet & 5 points, 1 level & 0.19 & 0.19 & 0.19 & 0.47 & 0.49 & 0.48 & 0.66 & 0.68 & 0.67 & 0.72 & 0.74 & 0.73 & 0.75 & 0.77 & 0.75 & 0.81 & 0.88 \\
\cline{2-22} \cline{3-22} \cline{4-22}
 & Original attention & Synthetic + real & Fréchet & 5 points, 4 levels & 0.13 & 0.14 & 0.14 & 0.42 & 0.42 & 0.42 & 0.68 & 0.69 & 0.68 & 0.77 & 0.78 & 0.77 & 0.82 & 0.82 & 0.82 & 0.82 & 0.89 \\
\cline{1-22} \cline{2-22} \cline{3-22} \cline{4-22}
\multirow[c]{4}{*}{practical} & \textbf{Main} & \textbf{Synthetic + real} & \textbf{Fréchet} & \textbf{5 points, 4 levels} & \textbf{0.12} & \textbf{0.12} & \textbf{0.12} & \textbf{0.42} & \textbf{0.40} & \textbf{0.41} & \textbf{0.66} & \textbf{0.63} & \textbf{0.64} & \textbf{0.74} & \textbf{0.71} & \textbf{0.72} & \textbf{0.77} & \textbf{0.74} & \textbf{0.75} & \textbf{0.80} & \textbf{0.86} \\
\cline{2-22} \cline{3-22} \cline{4-22}
 & \multirow[c]{3}{*}{Training strategy} & Synthetic only & Fréchet & 5 points, 4 levels & 0.00 & 0.00 & 0.00 & 0.00 & 0.00 & 0.00 & 0.00 & 0.00 & 0.00 & 0.00 & 0.00 & 0.00 & 0.00 & 0.00 & 0.00 & 0.05 & 0.05 \\
\cline{3-22} \cline{4-22}
 &  & Real only & Fréchet & 5 points, 4 levels & 0.00 & 0.00 & 0.00 & 0.00 & 0.00 & 0.00 & 0.06 & 0.04 & 0.05 & 0.13 & 0.09 & 0.11 & 0.22 & 0.16 & 0.18 & 0.45 & 0.56 \\
\cline{3-22} \cline{4-22}
 &  & Real only (25k) & Fréchet & 5 points, 4 levels & 0.06 & 0.06 & 0.06 & 0.27 & 0.23 & 0.25 & 0.53 & 0.45 & 0.48 & 0.62 & 0.54 & 0.57 & 0.69 & 0.59 & 0.63 & 0.68 & 0.74 \\
\cline{1-22} \cline{2-22} \cline{3-22} \cline{4-22}
\bottomrule
\end{tabular}%
}
\end{table*}
\begin{table*}[p]
\centering
\setlength{\tabcolsep}{3pt}
\caption{CryoTracer ablations on $\alpha$-synuclein. Precision (P), recall (R), and F1 are averaged over micrographs at each distance tolerance (in \AA{}). Box IoU and coverage measure spatial overlap. Synthetic pretraining uses up to 20,000 steps. Real-data training uses up to 5,000 steps unless marked 25k. Synthetic-only rows evaluate the same frozen checkpoint in both regimes. Bold labels identify the main model configuration. Bold scores mark the highest unrounded value per metric and regime. Higher scores are better. Training and architecture details appear in Appendix~\ref{sec:experiment-series}.}
\label{tab:asyn-ablation}
\resizebox{\textwidth}{!}{%
\begin{tabular}{lllllrrrrrrrrrrrrrrrrr}
\toprule
 &  &  &  &  & P@25 & R@25 & F1@25 & P@50 & R@50 & F1@50 & P@100 & R@100 & F1@100 & P@150 & R@150 & F1@150 & P@200 & R@200 & F1@200 & Box IoU & Coverage \\
Regime & Experiment & training & distance & architecture &  &  &  &  &  &  &  &  &  &  &  &  &  &  &  &  &  \\
\midrule
\multirow[c]{10}{*}{standard} & \textbf{Main} & \textbf{Synthetic + real} & \textbf{Fréchet} & \textbf{5 points, 4 levels} & 0.02 & 0.03 & 0.02 & \textbf{0.22} & 0.24 & \textbf{0.23} & \textbf{0.54} & \textbf{0.56} & \textbf{0.54} & \textbf{0.64} & 0.67 & 0.64 & \textbf{0.75} & \textbf{0.77} & \textbf{0.75} & \textbf{0.75} & 0.84 \\
\cline{2-22} \cline{3-22} \cline{4-22}
 & \multirow[c]{3}{*}{Training strategy} & Synthetic only & Fréchet & 5 points, 4 levels & 0.00 & 0.00 & 0.00 & 0.00 & 0.00 & 0.00 & 0.00 & 0.00 & 0.00 & 0.00 & 0.00 & 0.00 & 0.00 & 0.00 & 0.00 & 0.01 & 0.02 \\
\cline{3-22} \cline{4-22}
 &  & Real only & Fréchet & 5 points, 4 levels & 0.00 & 0.00 & 0.00 & 0.04 & 0.04 & 0.04 & 0.15 & 0.14 & 0.15 & 0.29 & 0.28 & 0.27 & 0.48 & 0.46 & 0.45 & 0.51 & 0.63 \\
\cline{3-22} \cline{4-22}
 &  & Real only (25k) & Fréchet & 5 points, 4 levels & 0.02 & 0.02 & 0.02 & 0.15 & 0.16 & 0.15 & 0.48 & 0.49 & 0.48 & 0.62 & 0.62 & 0.61 & 0.70 & 0.69 & 0.68 & 0.71 & 0.81 \\
\cline{2-22} \cline{3-22} \cline{4-22}
 & \multirow[c]{2}{*}{Distance metric} & Real only & Chamfer & 5 points, 4 levels & 0.00 & 0.00 & 0.00 & 0.00 & 0.00 & 0.00 & 0.00 & 0.00 & 0.00 & 0.00 & 0.00 & 0.00 & 0.00 & 0.00 & 0.00 & 0.35 & 0.48 \\
\cline{3-22} \cline{4-22}
 &  & Synthetic + real & Chamfer & 5 points, 4 levels & 0.00 & 0.00 & 0.00 & 0.00 & 0.00 & 0.00 & 0.01 & 0.01 & 0.01 & 0.01 & 0.01 & 0.01 & 0.02 & 0.02 & 0.02 & 0.69 & \textbf{0.87} \\
\cline{2-22} \cline{3-22} \cline{4-22}
 & \multirow[c]{2}{*}{Number of points} & \multirow[c]{2}{*}{Synthetic + real} & \multirow[c]{2}{*}{Fréchet} & 3 points, 4 levels & \textbf{0.04} & \textbf{0.05} & \textbf{0.04} & 0.21 & \textbf{0.25} & 0.22 & 0.47 & 0.56 & 0.50 & 0.59 & 0.68 & 0.61 & 0.66 & 0.74 & 0.68 & 0.73 & 0.83 \\
 &  &  &  & 10 points, 4 levels & 0.01 & 0.01 & 0.01 & 0.20 & 0.22 & 0.21 & 0.49 & 0.55 & 0.51 & 0.63 & \textbf{0.68} & \textbf{0.65} & 0.70 & 0.76 & 0.72 & 0.74 & 0.84 \\
\cline{2-22} \cline{3-22} \cline{4-22}
 & Feature pyramid & Synthetic + real & Fréchet & 5 points, 1 level & 0.01 & 0.01 & 0.01 & 0.10 & 0.12 & 0.10 & 0.43 & 0.47 & 0.44 & 0.58 & 0.64 & 0.59 & 0.66 & 0.73 & 0.68 & 0.73 & 0.83 \\
\cline{2-22} \cline{3-22} \cline{4-22}
 & Original attention & Synthetic + real & Fréchet & 5 points, 4 levels & 0.01 & 0.01 & 0.01 & 0.09 & 0.10 & 0.10 & 0.34 & 0.36 & 0.34 & 0.56 & 0.59 & 0.56 & 0.66 & 0.69 & 0.67 & 0.69 & 0.79 \\
\cline{1-22} \cline{2-22} \cline{3-22} \cline{4-22}
\multirow[c]{4}{*}{practical} & \textbf{Main} & \textbf{Synthetic + real} & \textbf{Fréchet} & \textbf{5 points, 4 levels} & \textbf{0.01} & \textbf{0.01} & \textbf{0.01} & \textbf{0.13} & \textbf{0.12} & \textbf{0.12} & \textbf{0.45} & \textbf{0.43} & \textbf{0.43} & \textbf{0.61} & \textbf{0.58} & \textbf{0.58} & \textbf{0.70} & \textbf{0.66} & \textbf{0.66} & \textbf{0.70} & \textbf{0.79} \\
\cline{2-22} \cline{3-22} \cline{4-22}
 & \multirow[c]{3}{*}{Training strategy} & Synthetic only & Fréchet & 5 points, 4 levels & 0.00 & 0.00 & 0.00 & 0.00 & 0.00 & 0.00 & 0.00 & 0.00 & 0.00 & 0.00 & 0.00 & 0.00 & 0.00 & 0.00 & 0.00 & 0.01 & 0.01 \\
\cline{3-22} \cline{4-22}
 &  & Real only & Fréchet & 5 points, 4 levels & 0.00 & 0.00 & 0.00 & 0.00 & 0.00 & 0.00 & 0.01 & 0.01 & 0.01 & 0.04 & 0.03 & 0.03 & 0.07 & 0.06 & 0.06 & 0.28 & 0.41 \\
\cline{3-22} \cline{4-22}
 &  & Real only (25k) & Fréchet & 5 points, 4 levels & 0.00 & 0.00 & 0.00 & 0.01 & 0.01 & 0.01 & 0.09 & 0.10 & 0.09 & 0.20 & 0.19 & 0.18 & 0.27 & 0.26 & 0.25 & 0.44 & 0.56 \\
\cline{1-22} \cline{2-22} \cline{3-22} \cline{4-22}
\bottomrule
\end{tabular}%
}
\end{table*}
\clearpage

\begin{figure*}[t]
    \centering
    \includegraphics[width=\textwidth]{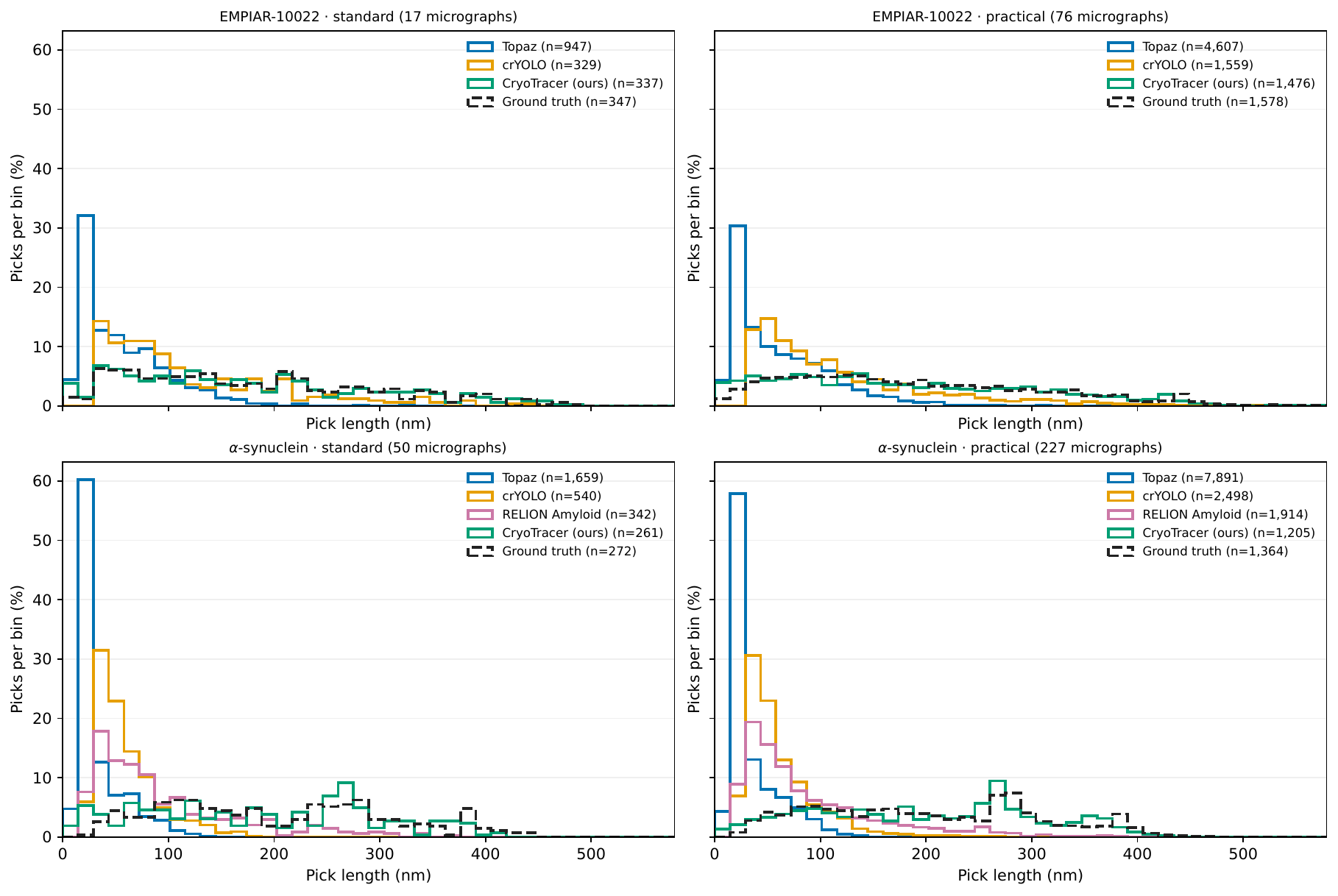}
    \caption{Picked-filament length distributions for EMPIAR-10022 (top) and $\alpha$-synuclein (bottom), under the standard (left) and practical (right) evaluation regimes. Length is measured along each polyline in nanometres. Histograms share 40 equal-width bins and show the percentage of each method's picks in each bin, summing to $100\%$ per method within a panel. The dashed curve denotes ground truth. Legend counts give the total number of picks.}
    \label{fig:pick-lengths}
\end{figure*}

\begin{figure*}[t]
    \centering
    \includegraphics[width=0.9\textwidth,height=0.82\textheight,keepaspectratio]{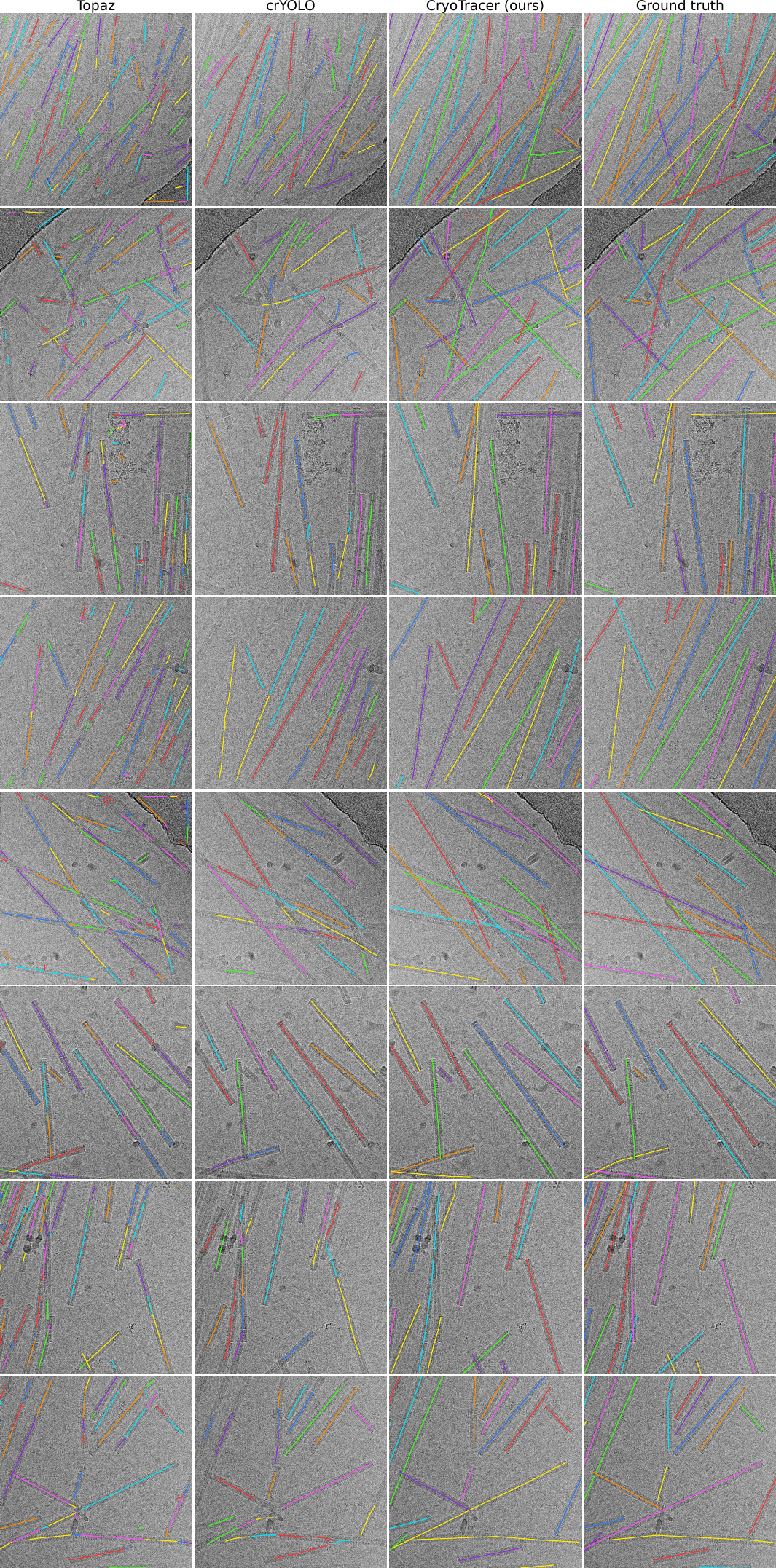}
    \caption{EMPIAR-10022, standard regime. Eight selected micrographs from the corresponding evaluation set, with method predictions and ground truth labeled in the panels. Images are low-pass filtered and contrast adjusted for display. Colors distinguish curves within a panel and do not indicate matched identities across methods.}
    \label{fig:empiar10022-standard-predictions}
\end{figure*}

\begin{figure*}[t]
    \centering
    \includegraphics[width=0.9\textwidth,height=0.82\textheight,keepaspectratio]{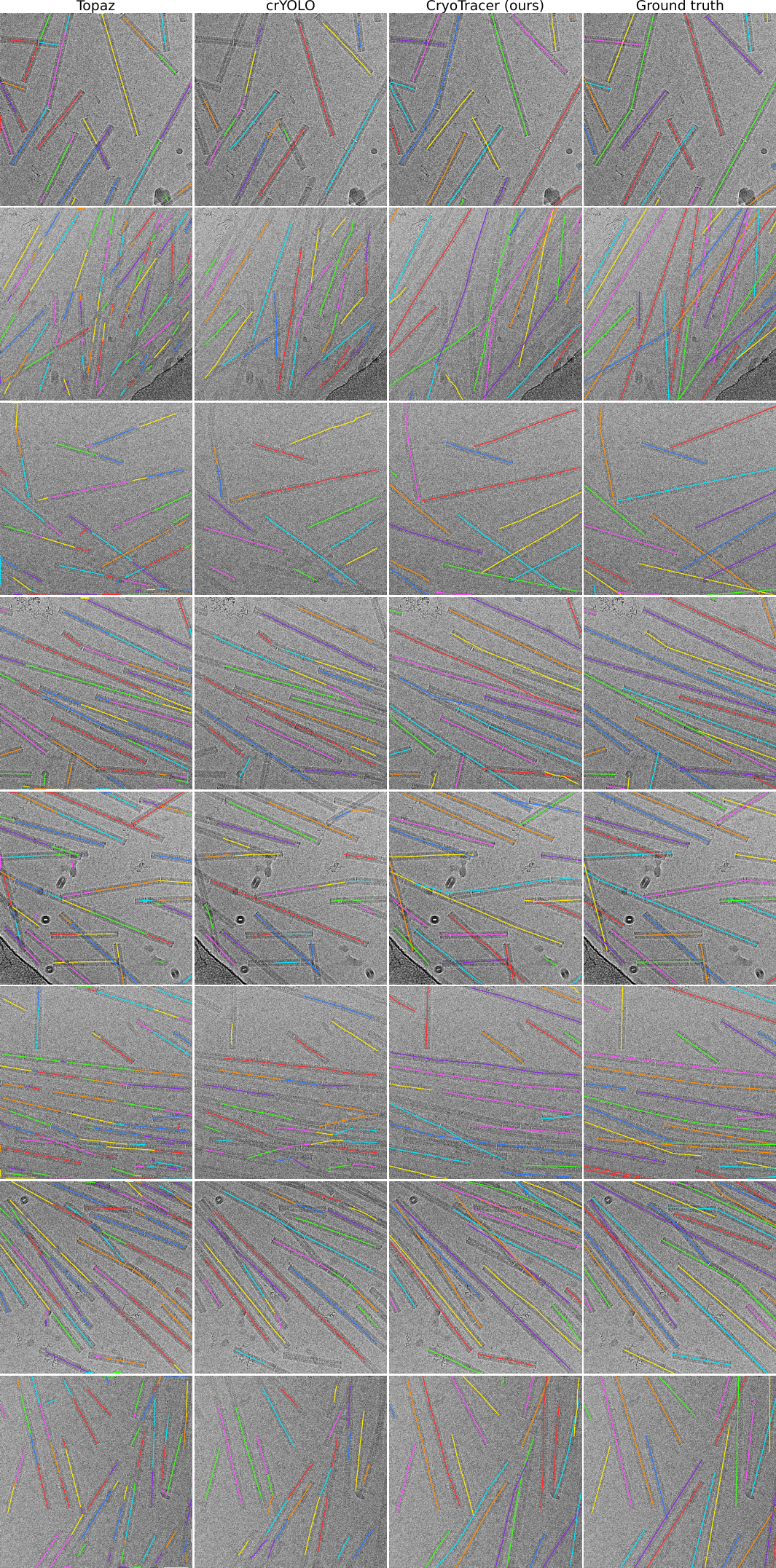}
    \caption{EMPIAR-10022, practical regime. Eight selected micrographs from the corresponding evaluation set, with method predictions and ground truth labeled in the panels. Images are low-pass filtered and contrast adjusted for display. Colors distinguish curves within a panel and do not indicate matched identities across methods.}
    \label{fig:empiar10022-practical-predictions}
\end{figure*}

\begin{figure*}[t]
    \centering
    \includegraphics[width=0.9\textwidth,height=0.82\textheight,keepaspectratio]{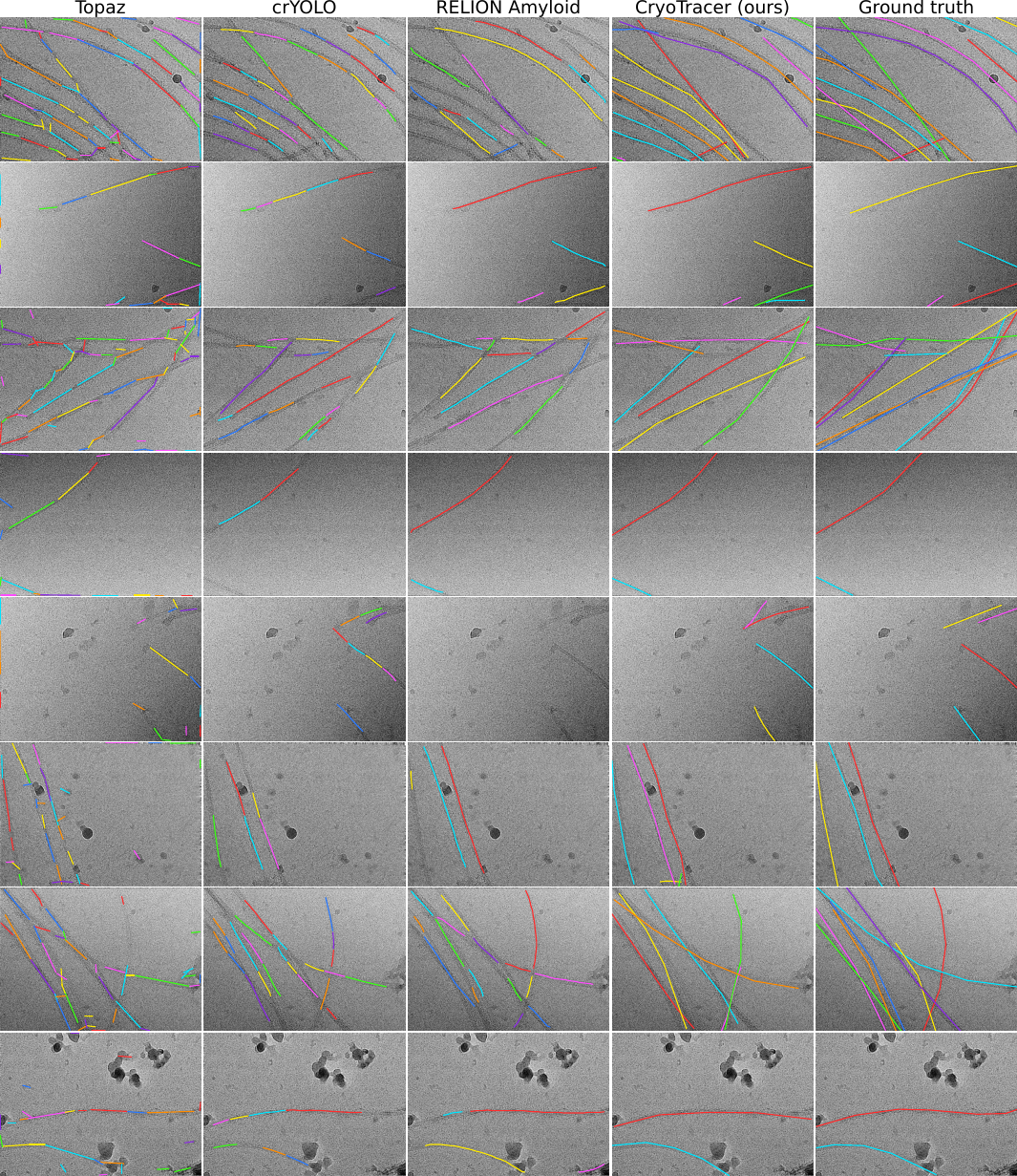}
    \caption{$\alpha$-synuclein, standard regime. Eight selected micrographs from the corresponding evaluation set, with method predictions and ground truth labeled in the panels. Images are low-pass filtered and contrast adjusted for display. Colors distinguish curves within a panel and do not indicate matched identities across methods.}
    \label{fig:asyn-standard-predictions}
\end{figure*}

\begin{figure*}[t]
    \centering
    \includegraphics[width=0.9\textwidth,height=0.82\textheight,keepaspectratio]{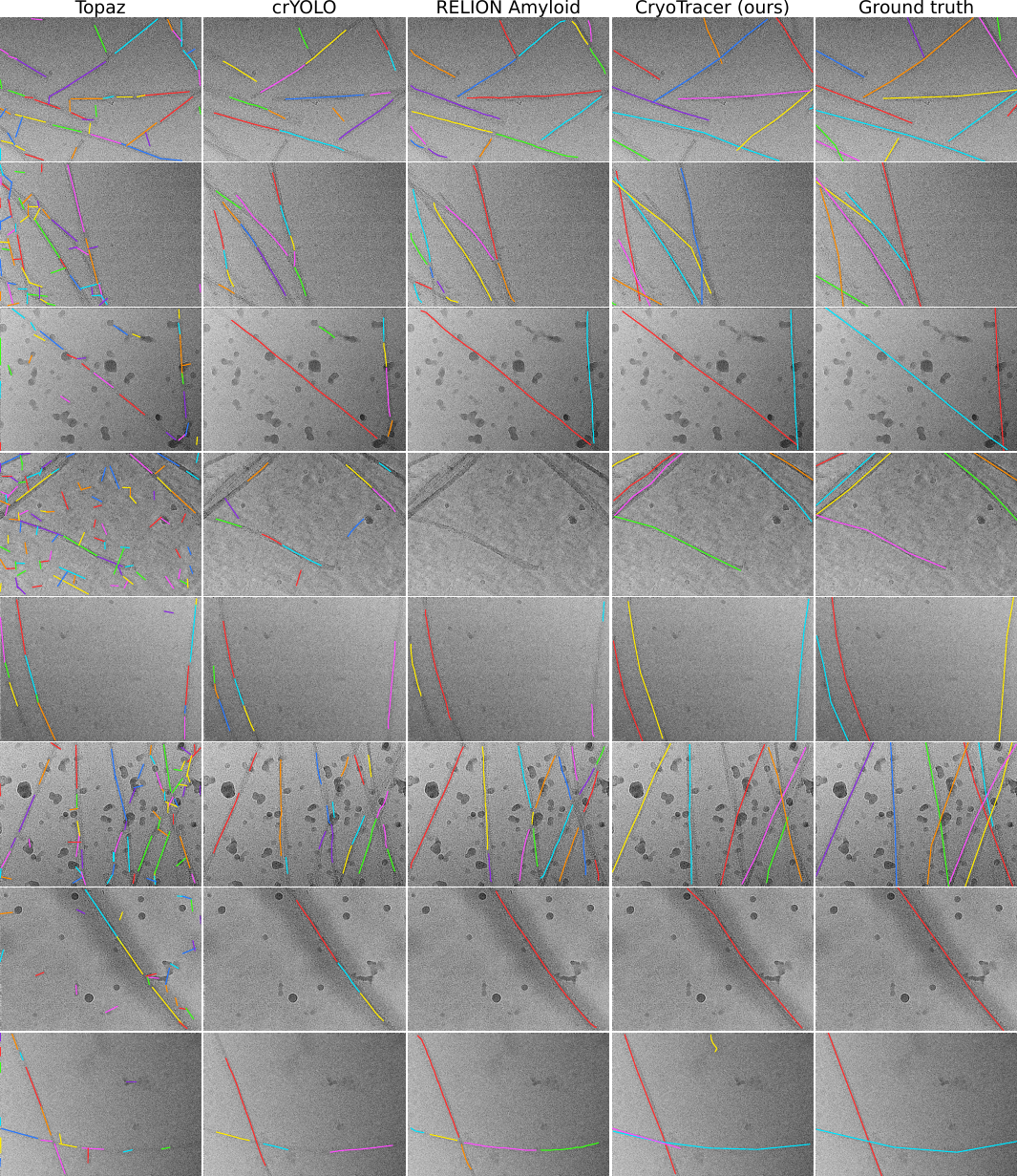}
    \caption{$\alpha$-synuclein, practical regime. Eight selected micrographs from the corresponding evaluation set, with method predictions and ground truth labeled in the panels. Images are low-pass filtered and contrast adjusted for display. Colors distinguish curves within a panel and do not indicate matched identities across methods.}
    \label{fig:asyn-practical-predictions}
\end{figure*}

\end{document}